# Artificial Intelligence Literacy and Sustainable Development: An Ethical Governance and Development Goals Framework

**Md. Masudul Islam** (Corresponding Author)
masudulislam11@gmail.com
ORCID: 0000-0001-7643-5420
Bangladesh University of Business and Technology, Dhaka, Bangladesh
Present Address: BUBT, Rupnagar RA, Mirpur-1216, Dhaka, Bangladesh

**Mirza Niaz Morshed**
mirza.n.morshed@gmail.com
ORCID: 0009-0005-7918-9604
Bangladesh University of Business and Technology, Dhaka, Bangladesh

**Md. Shafiqul Islam**
msislam@bubt.edu.bd
ORCID: 0000-0001-9469-2041
Bangladesh University of Business and Technology, Dhaka, Bangladesh

## Abstract
AI literacy provides foundational competencies that support ethical, transparent, and sustainable technological development, although higher-order capabilities such as governance, critical evaluation, and strategic decision-making extend beyond basic literacy into advanced levels of AI competency. This study positions AI literacy as a governance capacity that complements and strengthens all 17 SDGs. It introduces a six-level taxonomy of artificial intelligence reasoning and ethics that extends traditional learning models by incorporating ethical judgement and strategic foresight. This taxonomy forms the foundation of an integrated framework linking education, governance, and sustainable development. A survey of 300 participants from diverse professional backgrounds within a national context which reveals strong technical awareness but limited ethical and governance readiness, highlighting critical gaps in public capacity to manage artificial intelligence responsibly. Findings show that ethical reasoning and reflective thinking are the strongest predictors of sustainable and trustworthy artificial intelligence use. The study proposed to embed literacy-based competencies into curricula, institutional policies, and governance mechanisms to accelerate equitable and responsible progress toward sustainable development goals.

# 1. Introduction
Recent progress in artificial intelligence (AI) has rapidly changed economies (Trabelsi, 2024; Brandao, 2025), societies (Rashid and Kausik, 2024), and education (Rawas, 2024). AI can handle large amounts of data, create insights, automate tasks, and aid in making choices (Duan et al., 2019), which offers significant opportunities but also poses risks (Hutter and Hutter, 2021). AI's rising impact puts pressure on societies to adopt systems and develop the skills to understand, manage, and benefit from them responsibly, particularly in light of challenges such as algorithmic bias, privacy concerns, and inequality caused by automation (Khakurel et al., 2018; Khogali and Mekid, 2023). As AI continues to reshape society, understanding AI is no longer optional but essential for education and responsible citizenship (Alexandre et al., 2021). This understanding begins with foundational AI literacy encompassing basic awareness and use of AI systems and progressively develops into higher-order competencies such as critical evaluation, ethical reasoning, and responsible governance. (Bostrom and Yudkowsky, 2018; Cañas, 2022). Education and government leaders increasingly recognize that technical knowledge alone is insufficient (Muthukrishna et al., 2025). Instead, AI literacy is becoming crucial for democratic participation, ethical decision-making, and socially responsible innovation (Hristovska, 2023; Holmes et al., 2023). At the same time, the global community is guided by the United Nations Sustainable Development Goals (SDGs), a set of seventeen interconnected targets aimed at achieving peace, prosperity, and environmental sustainability by 2030 (UNDP, 2025). AI presents both opportunities and challenges for achieving these goals (Vinuesa et al., 2020). It accelerates progress in education (SDG 4) through personalized learning (Artyukhov et al., 2024), supports climate action (SDG 13) through predictive modelling (Al-Raeei, 2024), and strengthens institutions (SDG 16) by enhancing transparency and accountability (UN Global Compact, 2025). However, without equitable access and ethical understanding, AI can also exacerbate inequality, bias, and environmental harm (Pendyala, 2024). Therefore, the central issue is not whether AI can contribute to the SDGs (Jasper, 2024), but whether societies possess the capacity to govern and align AI with sustainable development objectives. Despite growing attention to AI literacy, existing research primarily treats it as an educational or technical concept rather than a governance mechanism. Most studies focus on domain-specific applications such as climate or education (Filho et al., 2025; Cebesoy and Önger, 2025), without recognizing AI literacy as an integrative capability that ensures ethical alignment across all seventeen SDGs. Furthermore, limited research integrates educational frameworks, governance strategies, and sustainability policies into a unified model. This gap restricts the transformation of AI literacy from a classroom-based concept into a systemic driver of sustainable development. To address this limitation, this study reconceptualizes AI literacy as a governance-oriented capacity that integrates education, ethics, and sustainability into a unified framework. A central contribution is the development of the Artificial Intelligence Reasoning and Ethics (AIRE) Taxonomy, a six-level model that extends Bloom's learning hierarchy by incorporating ethical reasoning and strategic foresight. The AIRE Taxonomy provides a structured pathway through which AI literacy evolves from basic understanding to institutional and policy-level governance, linking learning processes with sustainable decision-making. Building on this foundation, the study proposes the AI–SDG Nexus Framework, which systematically maps AI literacy competencies to all 17 Sustainable Development Goals. The research adopts a conceptual–analytical design supported by an empirical survey of 300 participants from diverse professional backgrounds. The findings reveal that while technical awareness of AI is relatively strong, ethical reasoning and governance literacy remain limited. Importantly, the results indicate that AI literacy enables governance by supporting ethical and informed decision-making capacity conceptualized here as an "18$^{th}$ SDG" heuristic that integrates education, governance, and sustainability to support progress across all 17 goals. Accordingly, this study aims to:

- Define AI literacy as a governance skill that integrates education, ethics, and sustainability throughout all 17 SDGs.
- Create and test the AIRE Taxonomy as a framework for teaching and governance, connecting thinking, ethical, and civic skills.
- Build and display the AI–SDG Nexus, showing how literacy areas match SDG groups and governance actions.
- Evaluate AI literacy preparedness and perceived SDG connections across various job fields.
- Suggest practical policy ideas for including literacy in national governance, institutional plans, and sustainable development efforts.

The remainder of this paper is organized as follows. Section 2 reviews the literature on AI literacy and its relationship with sustainable development. Section 3 presents the methodological framework. Section 4 introduces the theoretical foundations and the AIRE Taxonomy. Section 5 develops the AI–SDG Nexus and mapping analysis. Section 6 discusses governance and policy implications, followed by empirical validation in Section 7. The final section concludes the study with key contributions, limitations, and directions for future research.

# 2. Literature Review

## 2.1. AI Literacy: From Educational Foundations to Governance Relevance

AI literacy has increasingly been recognized as a fundamental competency for individuals navigating an algorithmically mediated society (Al-Abdullatif, 2025). In this study, AI literacy refers strictly to foundational competencies (Recognize–Apply), while higher-order capabilities (Analyze–Govern) are treated as advanced competencies or expertise. Evolving from preceding paradigms such as digital literacy, data literacy, and computational thinking (Yue Yim, 2024), AI literacy extends beyond basic operational use of technology to include essential understanding of how AI systems function and influence individuals and society; however, deeper analytical reasoning, critical evaluation, and governance capabilities are more appropriately understood as advanced competencies that build upon foundational AI literacy rather than constituting literacy itself (Long & Magerko, 2020; Ng et al., 2021). Consequently, AI literacy integrates both the technical dimension encompassing knowledge of how models process data, learn patterns, and generate results and the socio-ethical dimension, which entails critical interpretation, ethical awareness, and reflective reasoning (Biagini, 2025) about the broader implications of AI technologies.

Contemporary scholarship conceptualizes AI literacy as a multidimensional construct that encompasses technical, cognitive, and ethical proficiencies (Ateşgöz, 2025). Zhou et al. (Zhou et al., 2022) identify four recurrent pillars within the literature: knowing and understanding, using and applying, evaluating and creating, and developing ethical awareness. Within higher education, Educause (Kassorla et al., 2024) positions AI literacy as a comprehensive civic and professional competency rather than an ancillary digital skill, emphasizing the integration of analytical reasoning, ethical evaluation, and creative innovation. Empirical studies further indicate that effective AI literacy development requires a balanced synthesis of algorithmic comprehension, responsible tool utilization, data ethics, and social consciousness (Wiese et al., 2025, Heung and Su, 2025).

Despite these advancements, the field lacks a universally accepted definition or standardized framework for assessing AI literacy (Eyal, 2025, Chee et al., 2024). The absence of shared benchmarks constrains comparative research across institutions and impedes coherent policy development (Huggins, 2009). As illustrated in **Figure 1**, the progression from traditional digital and data literacies toward AI literacy signifies a paradigmatic shift from operational engagement with technological tools to a more holistic, cognitive, and ethically informed understanding of intelligent systems.

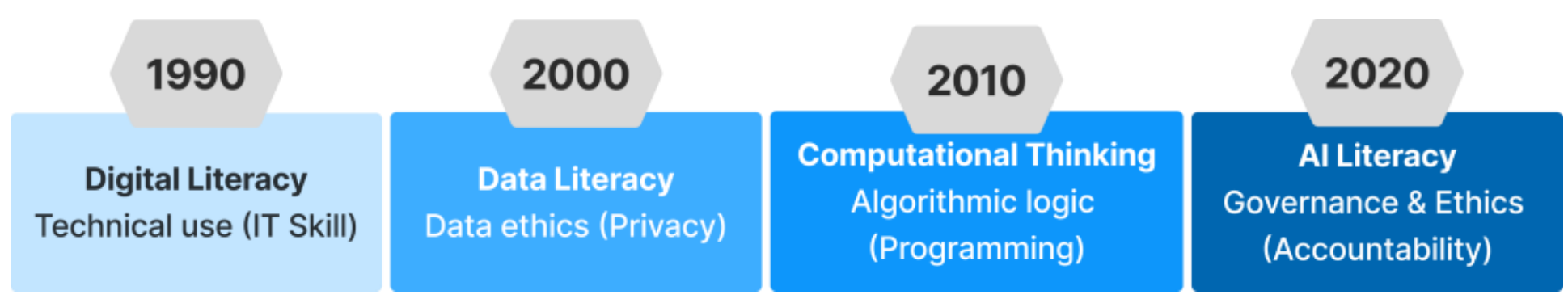

**Figure 1.** Evolution of Literacy Paradigms Toward AI Literacy. A conceptual timeline illustrating the paradigm shift in technological literacies from 1990 to 2020. The timeline is illustrative rather than historically discrete. The 10-year intervals represent the approximate decades during which each specific literacy became a primary focus in mainstream educational frameworks and societal discourse, rather than strict historical start-and-end dates.

AI literacy is increasingly recognized as both an essential educational foundation and a critical component of effective governance (Tadimalla and Maher, 2025). Empirical studies demonstrate that students' engagement, self-efficacy, and ethical awareness significantly influence learning outcomes (Miao et al., 2025; Kong et al.,2024). Evidence from Northern Cyprus indicates that AI literacy enhances inclusion and well-being, directly contributing to SDG 3 (Good Health and Well-Being) and SDG 4 (Quality Education) (Asim, 2025). Likewise, research from China underscores the potential of national education policies to institutionalize AI literacy as a catalyst for equitable digital transformation (Zhang et al., 2024).

From a governance standpoint, AI literacy is emerging as a cornerstone of risk management and public accountability (Tsarouhas and Grigoriadis, 2025). International and national policy frameworks, including those of UNESCO (UNESCO, 2023) and the OECD (OECD, 2024), emphasize capacity-building initiatives designed to equip not only technical specialists but also policymakers, educators, and citizens to engage with AI in an informed and responsible manner. Nonetheless, prevailing regulatory approaches remain largely compliance-oriented, prioritizing technical control over the cultivation of public understanding (Bernardo et al., 2024, Parker, 2000). In the absence of policymakers and citizens capable of critically interpreting algorithmic systems, ethical AI implementation and SDG-aligned innovation risk remaining aspirational goals (Floridi and Cowls, 2019). Accordingly, AI literacy functions at the intersection of education and governance it is both the pedagogical foundation that enables understanding of AI systems and the civic competence that ensures their alignment with human and planetary well-being (Otero et al., 2023). This dual function provides the conceptual basis for the governance-oriented framework developed in the subhesequent sections.
Recent research further reinforces the distinction between literacy and intelligence in AI systems themselves. For example, Raman et al. (Raman et al., 2024) demonstrated that systems such as ChatGPT can exhibit functional literacy in Sustainable Development Goal (SDG)-related tasks without possessing true contextual intelligence. This distinction highlights the importance of maintaining conceptual clarity between basic literacy, functional capability, and advanced reasoning, both in human and AI contexts. Building on this distinction, the present study does not equate AI literacy with expertise. Instead, it conceptualizes AI literacy as the foundational layer of a broader competency spectrum, which progressively develops into ethical reasoning, critical evaluation, and governance capability through the AIRE taxonomy. In this way, the framework preserves the conventional meaning of literacy while extending it into a structured developmental pathway.

### 2.2. Knowledge Management in AI-Driven Governance

The increasing integration of artificial intelligence into governance systems necessitates a parallel evolution in knowledge management (KM) practices. Knowledge management provides the structural foundation through which individuals and institutions acquire, organize, and apply knowledge for decision-making, policy formulation, and innovation. In governance contexts, KM ensures that data, information, and insights are transformed into actionable knowledge that supports transparency, accountability, and sustainable development. A useful conceptual lens within KM is the differentiation between levels of knowledge—know-what, know-how, and know-why. "Know-what" refers to basic awareness and factual understanding of AI technologies and their applications. "Know-how" reflects the ability to apply AI tools effectively in practical contexts, while "know-why" involves deeper analytical and critical understanding of underlying mechanisms, implications, and ethical considerations. These distinctions are particularly important in governance settings, where different stakeholders require varying depths of knowledge depending on their roles and responsibilities. In multi-level governance systems, knowledge requirements are inherently hierarchical and context-dependent. At the micro level, citizens and general users require foundational "know-what" competencies to engage with AI-enabled services. At the meso level, professionals and institutional actors require "know-how" to implement, evaluate, and manage AI systems. At the macro level, policymakers and regulators must possess "know-why" knowledge to design ethical frameworks, anticipate risks, and align AI deployment with societal and sustainability goals. This stratification highlights that AI-related competencies are not uniform but must be aligned with functional roles and governance structures. In the context of the Sustainable Development Goals (SDGs), knowledge management plays a critical enabling role by facilitating cross-sector collaboration, data sharing, and evidence-based policy design. AI systems amplify these processes by generating predictive insights and optimizing resource allocation; however, without effective KM structures, such capabilities remain underutilized or may lead to unintended consequences such as bias, misinterpretation, or policy misalignment. Despite its importance, the intersection of knowledge management, AI governance, and sustainable development remains underexplored. Existing approaches often emphasize technological capabilities without adequately addressing how knowledge is structured, distributed, and applied across governance systems. This gap underscores the need for frameworks that integrate knowledge hierarchies with AI competencies. Building on this perspective, AI literacy can be more precisely understood as the foundational layer of knowledge (know-what) within a broader knowledge management continuum, which progressively extends to applied competencies (know-how) and strategic understanding (know-why). The higher tiers of the AIRE taxonomy (**Section 4.1**) align closely with "know-how" and "know-why" dimensions in knowledge management frameworks. This alignment provides a clearer conceptual boundary between literacy and expertise, while supporting the development of structured competency frameworks such as the AIRE taxonomy.

### 2.3. Artificial Intelligence and the SDGs

AI is increasingly recognized as both a catalyst and a challenge in realizing the United Nations 2030 Sustainable Development Agenda (Arora and Mishra, 2019). The 17 Sustainable Development Goals (SDGs), as articulated by the United Nations, constitute an integrated framework for advancing global peace, prosperity, and environmental sustainability. AI's transformative capacity are encompassing predictive health analytics, precision agriculture, climate modelling, and transparent governance directly intersects with multiple SDGs (Vinuesa et al., 2020).

Existing research, however, often examines AI's contributions in isolation, focusing on individual goals rather than its systemic interdependencies across the SDG framework. For instance, AI-enabled personalized learning advances SDG 4 (Quality Education), climate prediction technologies support SDG 13 (Climate Action), and algorithmic oversight enhances SDG 16 (Peace, Justice, and Strong Institutions) (Maghsoudi et al., 2025, P.D.E., 2025, Herath et al., 2024). Nonetheless, limited attention has been paid to the mediating role of AI literacy in optimizing these impacts while mitigating associated risks such as algorithmic bias, social inequality, and misinformation (Jobin et al., 2019; Dwivedi et al., 2023). **Figure 2** presents a conceptual interpretation of the relationship between AI and the Sustainable Development Goals (SDGs), based on a synthesis of existing literature on AI applications rather than explicit references to "AI literacy" alone. While direct studies linking AI literacy to SDGs remain limited, substantial research exists on individual components of AI capabilities, including machine learning applications in climate science, healthcare, agriculture, and governance. Accordingly, the classification reflects the extent to which different SDGs depend on data-driven decision-making, algorithmic systems, and governance mechanisms. Environmental goals (SDGs 13, 14, and 15), for instance, are increasingly supported by AI-driven climate modelling, biodiversity monitoring, and environmental intelligence systems, indicating a strong relationship with AI technologies. Therefore, the spectrum should be interpreted as a conceptual mapping of AI application intensity and governance relevance rather than a rigid or purely literature-count-based classification.

Indirect AI Engagement
(Human development & inclusion-focused goals)
[SDG: 1, 2, 5, 10]

Application-Oriented AI Integration
(Infrastructure, economy, and service optimization)
[SDG: 6, 7, 8, 9, 11, 12]

Data-Intensive & Governance-Driven AI Systems
(Education, environment, and institutional governance)
[SDG: 4, 13, 14, 15, 16, 17]

Mediating Capability: AI Literacy
(Foundational → Applied → Governance Competency)

Ethical & Governance Filter transforms raw AI impact into measurable, sustainable outcomes

**Figure 2.** AI–SDG relationship spectrum illustrating varying modes of AI engagement across SDG clusters. The classification is based on a conceptual synthesis of literature on AI applications rather than solely on the explicit use of the term "AI literacy." The spectrum reflects differences in data intensity, application complexity, and governance relevance across SDGs. AI literacy is positioned as a mediating capability that enables the translation of AI-driven outputs into ethical, sustainable, and policy-relevant outcomes.

### 2.4. Synthesis and Research Gap

The intersection of AI literacy and sustainable development exposes three enduring research gaps.

i. **Absence of comprehensive alignment across the 17 SDGs**: Existing scholarship largely examines fragmented connections most often within education, climate action, or health without establishing an integrative framework that systematically links AI literacy competencies to the full spectrum of Sustainable Development Goals.
ii. **Weak governance integration**: AI literacy is seldom conceptualized as a governance instrument capable of embedding ethical reasoning, regulatory compliance, and risk-management mechanisms within sustainability agendas.
iii. **Deficiency of evaluative metrics:** Empirical frameworks remain scarce in assessing how AI literacy influences sustainability outcomes or how governance systems operationalize literacy into measurable SDG progress.

### 2.5. Toward an Integrated AI-SDG Nexus

Emerging scholarship increasingly recognizes AI literacy as a pivotal integrative construct bridging human capability, governance, and sustainability. When aligned with the multidimensional objectives of the Sustainable Development Goals (SDGs), the core competencies of AI literacy technical, ethical, and socio-civic can collectively function as a governance-driven capacity, effectively conceptualized as a heuristic "18th Goal" or enabling governance metaphor, representing a cross-cutting capacity that supports alignment across all SDGs rather than a formally defined additional goal (Nunes and Nunes, 2024). This meta-competency would enable coherent progress across all seventeen SDGs by embedding ethical reasoning, responsible innovation, and civic engagement into technological and institutional practices. As illustrated in **Figure 3**, AI literacy is positioned at the centre of the SDG clusters—People, Planet, Prosperity, Peace, and Partnership—highlighting its cross-cutting influence and integrative potential. For instance:

1. SDG 1 (No Poverty): AI literacy promotes inclusive financial technologies and equitable data utilization.

2. SDG 5 (Gender Equality): Awareness of algorithmic bias fosters fairness and representation in AI systems.
3. SDG 9 (Industry, Innovation, and Infrastructure): Innovation literacy supports sustainable and ethical AI deployment.
4. SDG 13 (Climate Action): Competence in interpreting AI-driven climate models enhances ethical environmental decision-making.
5. SDG 16 (Peace, Justice, and Strong Institutions): Civic literacy strengthens transparency and algorithmic accountability in governance.

This mapping reframes AI literacy not as a discrete educational outcome but as an enabling framework that underpins ethical decision-making, institutional coherence, and societal resilience across all SDGs. Building on these insights, the present study contributes by:

i. Developing a governance-oriented conceptual framework that situates AI literacy as an enabler of all seventeen SDGs.
ii. Constructing a systematic mapping between AI literacy competencies and SDG domains to address the gap in comprehensive alignment.
iii. Proposing measurable indicators and policy pathways linking AI literacy with ethical and sustainable governance outcomes.
iv. Framing AI literacy as a meta-goal that unites education, policy, and societal transformation within the context of the 2030 Agenda.

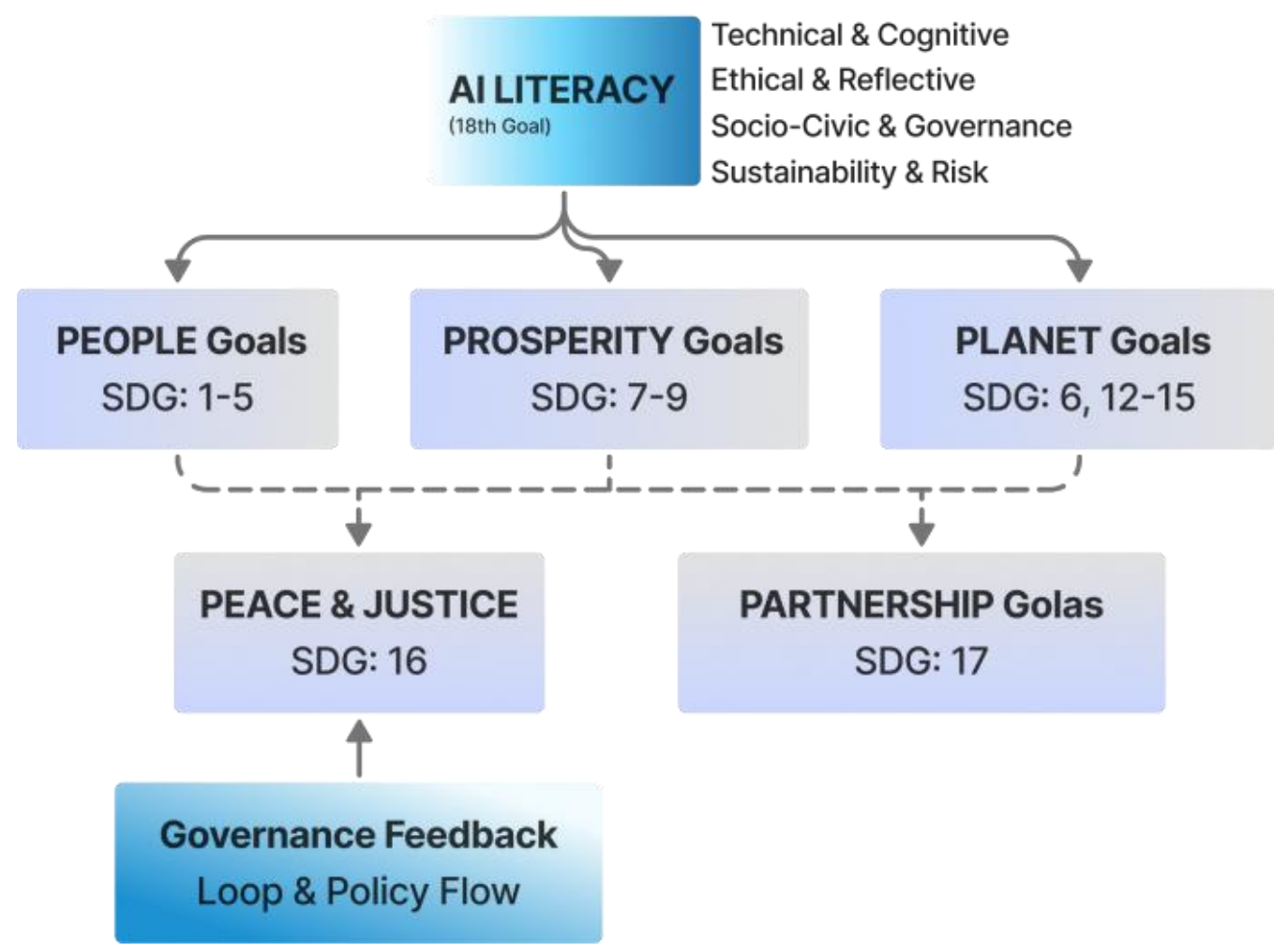

**Figure 3.** Conceptual overview of the AI–SDG Nexus positioning AI literacy

The diagram organizes the 17 Sustainable Development Goals into thematic clusters—People (SDG 1–5), Planet (SDG 6, 12–15), Prosperity (SDG 7–9), Peace (SDG 16), and Partnership (SDG 17) and places AI literacy at the centre as the enabling meta-competency that connects and sustains them.
The AI–SDG Nexus metaphor in **Figure 4** illustrates AI literacy as an integrative, cross-cutting governance capacity (conceptualized as an "18$^{th}$ SDG" heuristic) is a cognitive and ethical bridge that illuminates all 17 SDGs. It channels learning into governance and sustainability through the AIRE Taxonomy, forming a continuous loop from education to policy action.

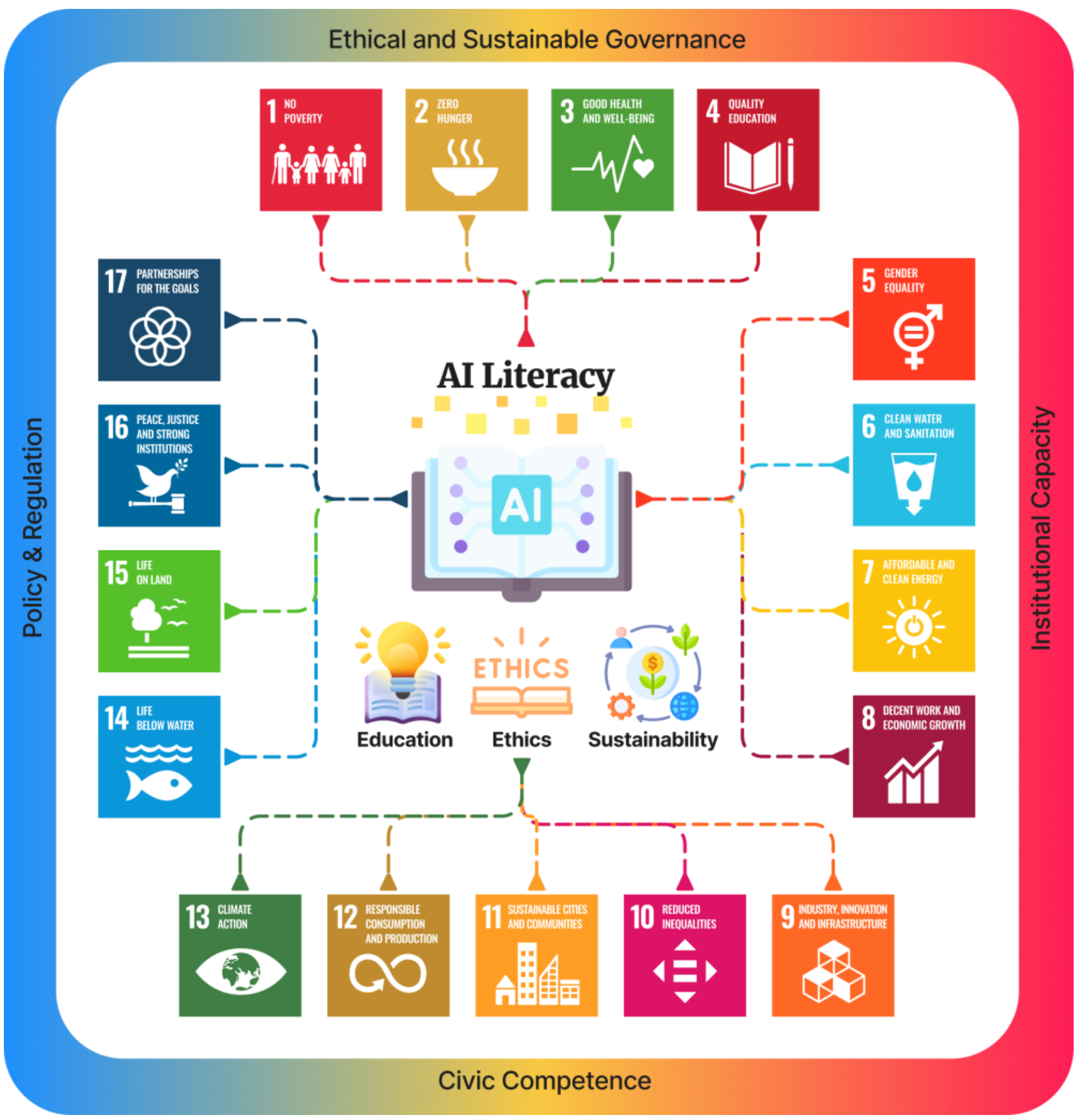

**Figure 4.** The AI–SDG Nexus Framework at a glance

# 3. Methodological Framework

A conceptual–analytical methodology was employed to integrate insights from the traditionally fragmented domains of education, governance, ethics, and sustainable development, each addressing AI literacy from distinct perspectives. Whereas technical research prioritizes algorithmic proficiency and educational studies focus on pedagogy, limited frameworks synthesize these dimensions with governance and policy analysis. The chosen design facilitates a systematic consolidation of theoretical, policy, and empirical evidence into a unified, governance-oriented model. In doing so, it positions AI literacy not merely as an educational competence but as a structural capacity underpinning ethical regulation and sustainable innovation within the SDG framework (AlSagri and Sohail, 2024).

## 3.1. Research Design

This study employs a conceptual–analytical research design that integrates systematic literature synthesis, policy document analysis, and framework development. Instead of empirical experimentation, the approach draws on interdisciplinary evidence spanning educational theory, AI ethics, sustainability governance, and policy studies to construct a governance-oriented model of AI literacy aligned with all 17 SDGs. The design follows a deductive–integrative logic synthesizing existing knowledge, identifying structural gaps, and formulating a unifying framework that links education, governance, and sustainability. The overall workflow, illustrated in **Figure 5**, depicts the integration of literature synthesis, policy review, and framework construction within the conceptual–analytical design.

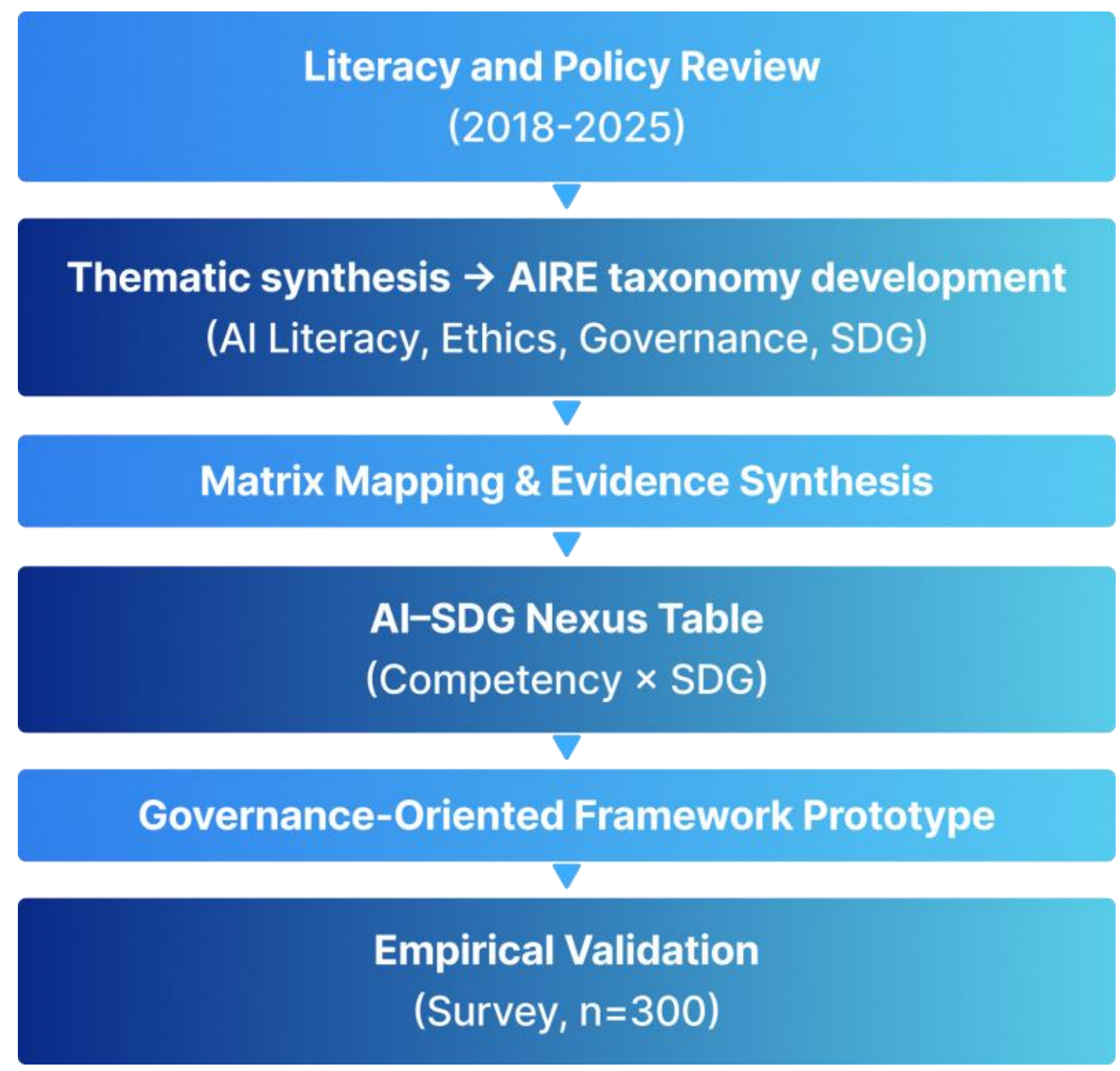

**Figure 5.** Conceptual–Analytical Research Design Workflow

The empirical component of the study is based on primary data collected through a voluntary and anonymous survey of 300 participants, ensuring alignment between the conceptual framework and real-world validation.

### 3.2. Conceptual Basis and Framework Logic

The framework integrates the PICO logic model (Population–Interest–Context) (Amir-Behghadami and Janati, 2020) with SMART-aligned objectives to ensure conceptual precision and policy relevance.

- **Population/Problem:** Learners, educators, policymakers, and institutions engaged in shaping AI education and governance ecosystems.
- **Interest/Intervention:** The formulation and implementation of AI literacy as a governance mechanism advancing the Sustainable Development Goals (SDGs).
- **Context:** The global sustainable development agenda outlined in the UN 2030 Agenda.

The integration of the SMART criteria—Specific, Measurable, Achievable, Relevant, and Time-bound (Bjerke and Renger, 2017) which ensures that the framework transcends conceptual alignment and attains operational clarity. By defining explicit indicators under each dimension, the model establishes specific learning and governance outcomes, measurable literacy benchmarks, achievable institutional goals, and relevance to national and global policy priorities. The time-bound element anchors these targets within the 2030 SDG horizon, reinforcing accountability and progress tracking. Consequently, the SMART integration transforms AI literacy from a theoretical construct into an actionable and evaluable mechanism for advancing sustainable governance.

### 3.3. Data Sources and Analytical Process

The study integrates evidence from three complementary sources.

- Peer-reviewed studies published between 2018 and 2025 on AI literacy, ethics education, governance frameworks, and AI applications in sustainability were systematically retrieved from Scopus, Web of Science, ScienceDirect, and UNESCO databases using the keywords AI literacy, governance, sustainable development, education, and SDG.
- Key frameworks and guidelines were examined from UNESCO, OECD, the UN AI and SDG Lab (Jarzebski et al., 2023), and selected national AI strategies, including the EU AI Act drafts and Singapore's Model AI Governance Framework.

- Extracted materials were coded under four analytic themes—(i) AI literacy domains, (ii) governance levers, (iii) ethical principles, and (iv) SDG targets—using iterative matrix mapping to identify intersections between competencies and sustainability goals.

The analytical process comprised three sequential stages:

- **Extraction**: Identification of recurring AI literacy competencies and governance mechanisms.
- **Synthesis**: Clustering of competencies into four domains—technical, ethical, civic, and risk governance.
- **Mapping**: Alignment of each competency with relevant SDG targets through relevance analysis, policy linkage, and indicator-based validation, resulting in the AI Literacy × SDG Mapping presented in **Table 4**.

### 3.3.1. Literature Search Strategy and Selection Criteria

To enhance methodological transparency and reproducibility, the literature synthesis followed a structured and replicable search and screening process. Relevant studies were retrieved from four major databases: Scopus, Web of Science, ScienceDirect, and UNESCO repositories. The search was conducted using combinations of the following keywords and Boolean operators:

(“AI literacy” OR “artificial intelligence literacy”) AND (“governance” OR “policy” OR “ethics”) AND (“sustainable development” OR “SDG” OR “sustainability”)

The time window for inclusion was 2018–2025, reflecting the most recent developments in AI literacy, governance, and sustainability research. The inclusion criteria of these studies were:

- Focused on AI literacy, AI governance, ethics, or AI applications related to sustainability
- Were peer-reviewed journal articles, conference papers, or policy reports
- Were published in English
- Provided conceptual, empirical, or policy-relevant insights

The exclusion criteria were:

- Focused solely on technical AI model development without relevance to literacy or governance
- Lacked relevance to sustainability or SDG-related themes
- Were duplicates or non-scholarly sources (e.g., editorials, opinion pieces without analytical contribution)

Finally, during the screening and selection process, the initial search yielded approximately 173 records across databases. After removing duplicates, 51 studies remained for title and abstract screening. Following full-text review based on the inclusion criteria, a final set of 7 studies and policy documents was retained for synthesis and framework development. Following full-text assessment, 7 studies and policy documents were retained for final synthesis. The selection process was guided by principles of transparency and reproducibility and follows a PRISMA-inspired structured review approach and aligns with recent structured review methodologies in sustainability research (Priyanka et al., 2026)., although not conducted as a full systematic literature review. While 7 key studies were retained for deep synthesis, additional supporting literature informed the broader conceptual framing.

### 3.3.2 Statistical Analysis Plan

To ensure analytical consistency between the conceptual framework and empirical validation, the study employs a structured statistical analysis plan comprising three sequential stages: descriptive, correlational, and regression analysis. Descriptive statistics (means, standard deviations) were first computed to summarize domain-wise AI literacy readiness across the five constructs: Technical Literacy (TL), Ethical Literacy (EL), Governance Literacy (GL), Sustainability Literacy (SL), and Nexus Awareness Index (NAI). Correlation analysis (Pearson’s r) was then conducted to examine interrelationships among literacy domains and their association with SDG awareness (NAI), providing initial evidence of structural linkages within the AI–SDG Nexus. Finally, a multiple linear regression model was estimated to assess the predictive influence of AI literacy domains on nexus awareness. The dependent variable was Nexus Awareness Index (NAI), while independent variables included Governance Literacy (GL), Ethical Literacy (EL), and Technical Literacy (TL). Sustainability literacy was excluded due to multicollinearity considerations identified during preliminary diagnostics. The regression model is specified as: $(NAI = \beta_0 + \beta_1(GL) + \beta_2(EL) + \beta_3(TL) + \varepsilon)$

Assumptions of linear regression were assessed at a high level, including linearity, independence of observations, and absence of severe multicollinearity. Residual diagnostics indicated no major violations affecting interpretability.

### 3.4. Framework Development Procedure

The development process followed an iterative, evidence-based approach. First, a conceptual alignment was conducted in which each Sustainable Development Goal (SDG) was reviewed to identify the specific domains where AI literacy functions as an enabling or mediating factor, for instance, bias awareness in relation to SDG 5 (Gender Equality) and environmental data literacy for SDG 13 (Climate Action). Next, a governance layer integration was undertaken, analysing existing AI governance mechanisms such as transparency mandates, algorithmic audits, and risk-classification frameworks to position AI literacy as both an educational outcome and a governance instrument. Subsequently, indicator formulation involved adapting measurable parameters, including accessibility compliance, bias gaps, and traceability metrics, from established SDG indicators to construct an evaluative link between literacy initiatives and sustainable development outcomes. Finally, validation and refinement were achieved through cross-referencing the proposed competencies with UNESCO and OECD frameworks, ensuring policy coherence and global applicability.

### 3.5. Ethical and Quality Considerations

This study combines a conceptual–analytical framework with an empirical component involving human participants. The empirical validation presented in Section 7 is based on a cross-sectional survey of 300 respondents from diverse professional backgrounds. The data collected are primary (not secondary) and were obtained through voluntary participation. As the survey did not collect any personally identifiable or sensitive information and posed minimal risk to participants, formal Institutional Review Board (IRB) approval was not required in accordance with applicable institutional and national research guidelines. Participation was fully voluntary, and informed consent was obtained from all respondents prior to completing the questionnaire. Participants were clearly informed about the purpose of the study, the academic use of the data, and their right to withdraw at any stage without penalty. To ensure data protection and confidentiality, no identifying information was collected or stored. All responses were anonymized at the point of collection and analyzed in aggregated form. Data were securely stored and accessed only by the research team. For the conceptual component, ethical rigor was maintained through the use of publicly available academic and policy sources, ensuring transparency, neutrality, and appropriate citation practices. The overall validity of the study is strengthened through the triangulation of conceptual, policy, and empirical evidence.

### 3.6. Analytical Framework Outcome

The proposed methodology culminates in the AI–SDG Nexus Framework, a governance-oriented conceptual model that positions AI literacy as a meta-competency underpinning all 17 Sustainable Development Goals. The following sections elaborate this process, first by establishing the theoretical foundations of the framework and then by translating these conceptual competencies into a governance-ready taxonomy.

The outcome of this analytical process is the identification of structured competency dimensions and governance linkages, which are synthesized into the Artificial Intelligence Reasoning and Ethics (AIRE) Taxonomy. Specifically, recurring themes identified during the coding stage such as technical understanding, ethical awareness, governance capability, and sustainability considerations were systematically organized into a hierarchical progression of competencies. This progression reflects increasing levels of cognitive, ethical, and institutional engagement, forming the conceptual basis of the AIRE framework presented in the subsequent section.

# 4. Theoretical Foundations of AI Literacy

Building on the methodological framework and thematic analysis described in Section 3, this section presents the theoretical foundations of AI literacy and introduces the AIRE taxonomy as a synthesized outcome of the literature review, policy analysis, and competency mapping process. Artificial

Intelligence literacy stands at the crossroads of education, ethics, and governance. Its foundations draw from learning theories, socio-technical perspectives, and moral frameworks that together define how societies cultivate responsible intelligence. This section unfolds through four interconnected layers: the educational basis of AI literacy, its socio-technical and governance dimensions, its ethical and critical pedagogy, and its scaling across micro-, meso-, and macro-levels of governance.

### 4.1. Educational and Pedagogical Foundations of AI Literacy: The AIRE Taxonomy

AI literacy is rooted in contemporary learning theory, which conceptualizes knowledge as an active and socially mediated process. From a constructivist perspective, learners develop mental models of how algorithms perceive, classify, and make decisions rather than merely memorizing technical procedures. In this regard, AI literacy extends digital literacy by requiring not only technical understanding but also cognitive, ethical, and civic engagement with the implications of machine reasoning.

To operationalize this multidimensional learning process, the study introduces a six-tier framework the AIRE Taxonomy that extends Bloom's hierarchy (Kastberg, 2003) to encompass ethical synthesis and governance foresight. Each tier specifies a learning outcome, a governance competency, and a measurable indicator, enabling evaluation across three systemic levels: micro (learner), meso (institutional), and macro (policy). As illustrated in **Figure 6**, the AIRE Taxonomy depicts the pedagogical progression of AI literacy from foundational recognition to advanced governance capability. The six levels of the AIRE taxonomy were derived by clustering coded themes into progressively complex competency groups, aligning foundational knowledge with higher-order ethical reasoning and governance capabilities.

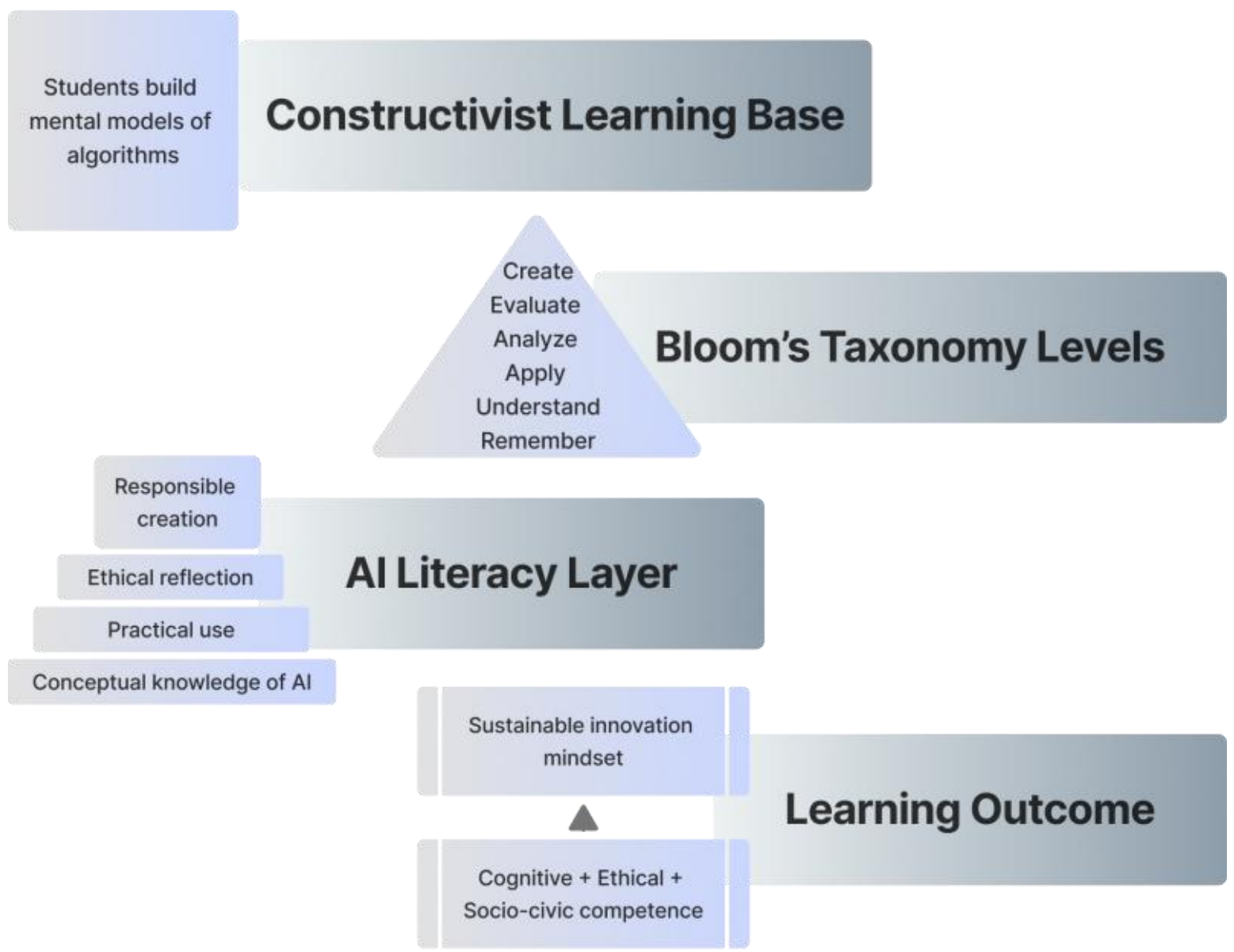

**Figure 6.** Pedagogical Progression of AI Literacy Competence

The AIRE Taxonomy functions simultaneously as a pedagogical framework and a policy evaluation instrument. It allows education ministries, accreditation agencies, and AI regulators to monitor measurable progress across learning ecosystems. Lower tiers (1–3) can be assessed through curriculum design and competency evaluations, while higher tiers (4–6) correspond to institutional audits, governance maturity indices, and the adoption of AI policies aligned with the Sustainable Development Goals (SDGs). By linking cognitive development with governance indicators, the AIRE framework transforms abstract ethical learning into evidence-based policy outcomes, ensuring that AI literacy evolves as both an educational and systemic governance capacity. **Table 1** summarizes the integration of the AIRE taxonomy across curriculum and governance dimensions.

**Table 1.** AIRE Taxonomy Table –Curriculum and Governance Integration

| AIRE Level | Learning Outcome | Governance Competency | Illustrative Indicator / Policy Metric |
| --- | --- | --- | --- |
| **1. Recognize –** Awareness | Identify AI concepts, common tools, and use cases. | Basic AI literacy inclusion in curriculum and training. | % of institutions offering AI awareness modules; literacy pre/post scores. |
| **2. Comprehend –** Understanding | Explain how AI systems function and affect society. | Interpretive understanding for informed participation. | Knowledge assessment scores; % of educators trained in AI fundamentals. |
| **3. Apply –** Responsible Use | Use AI tools ethically and effectively in learning or work. | Compliance with ethical guidelines and responsible innovation. | % of users trained under AI ethics modules; adherence to institutional AI policies. |
| **4. Analyze –** Critical Evaluation | Detect bias or misuse in datasets and model outputs. | Evidence-based accountability and risk assessment. | Number of algorithmic audits; frequency of bias reporting mechanisms. |
| **5. Integrate –** Ethical Synthesis | Design inclusive, transparent, and sustainable AI workflows. | Institutional integration of fairness and sustainability principles. | % of organizations adopting ethical AI frameworks; sustainability audit compliance. |
| **6. Govern –** Strategic Foresight | Translate literacy into policy, regulation, or institutional strategy. | Anticipatory governance and ethical foresight. | Presence of national AI literacy benchmarks; policy inclusion rate in SDG reports. |

**Figure 7** depicts the hierarchical structure of the AIRE taxonomy, extending Bloom's cognitive ladder into ethical and governance domains. The AIRE taxonomy was derived from the synthesis and mapping process described in **Section 3.3**, where recurring competency patterns were identified and structured into hierarchical levels. The AIRE framework thus provides the pedagogical and ethical scaffolding through which AI literacy develops from personal awareness to institutional accountability. It operationalizes the idea of learning to govern intelligence, linking classroom-level practice (Recognize → Apply) with systemic competencies (Integrate → Govern). This progression ensures that cognitive skill, ethical reflection, and civic responsibility evolve together forming the foundation for the socio-technical and governance analysis presented in the next section. The AIRE taxonomy conceptualizes AI literacy as a developmental continuum, where foundational literacy represents the entry level and progressively evolves into advanced competencies including ethical reasoning, critical analysis, and governance foresight.

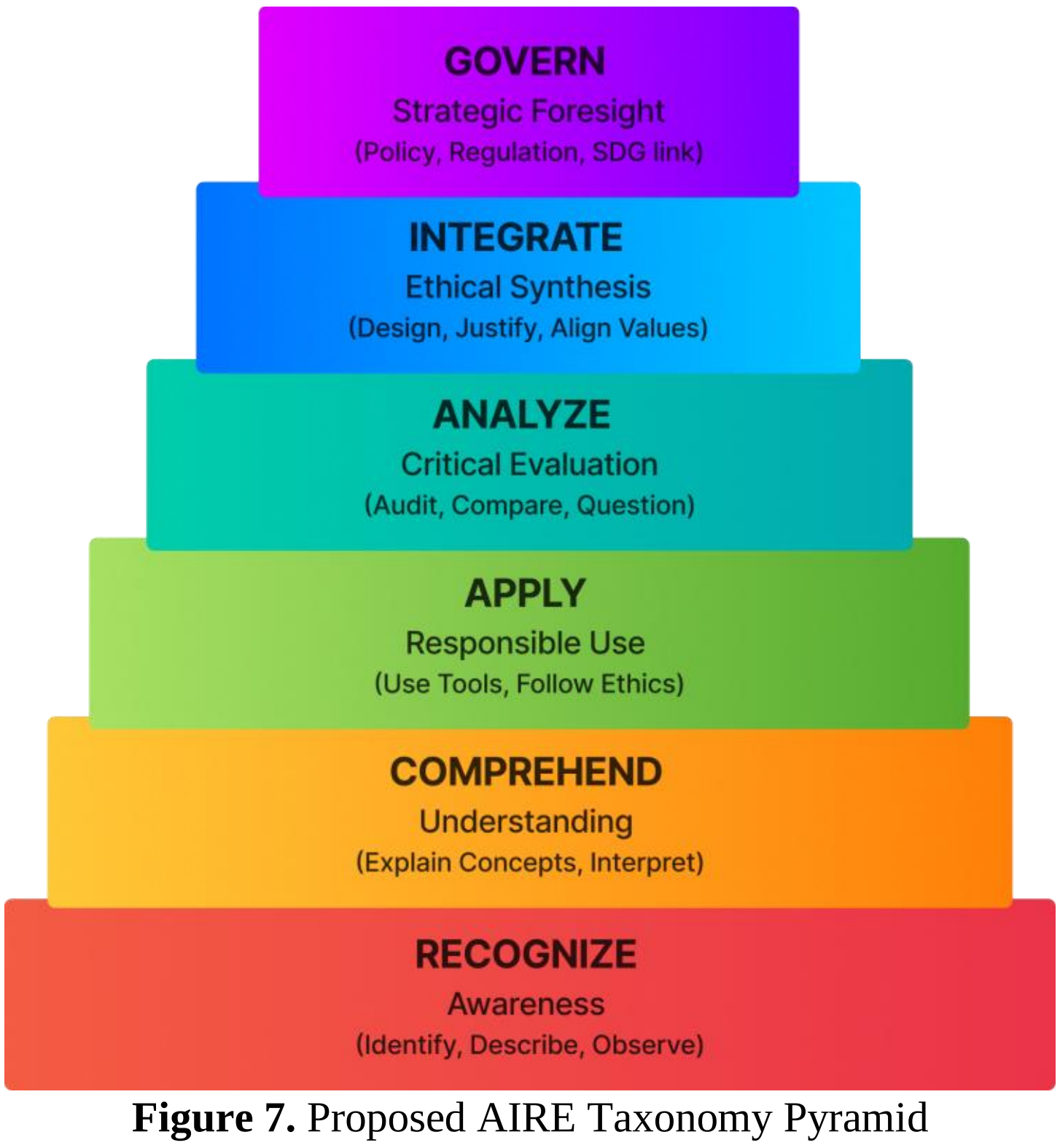

**Figure 7.** Proposed AIRE Taxonomy Pyramid

#### 4.1.1. Feasibility and Contextual Implementation of AIRE Taxonomy

While the AIRE taxonomy provides a comprehensive progression from foundational awareness to governance-level capability, its implementation must be understood within realistic educational and institutional constraints. In most academic contexts, particularly within undergraduate programs of two to three years, it is not feasible to achieve the full spectrum of competencies outlined in the higher tiers of the taxonomy. Accordingly, the AIRE framework should be interpreted as a multi-stage developmental model rather than a linear curriculum to be completed within a single program. Foundational levels: Recognize, Comprehend, and Apply which are correspond to AI literacy and can be effectively integrated into undergraduate and early-stage learning environments. In contrast, higher-order levels: Analyze, Integrate, and Govern which represent advanced competencies that align more closely with professional training, organizational learning, and policy-level engagement. This distinction reinforces the need to differentiate between AI literacy and AI expertise. While literacy encompasses essential knowledge and basic competencies, the upper tiers of the AIRE taxonomy extend into domains of critical evaluation, ethical synthesis, and governance foresight, which are more appropriately categorized as forms of expertise rather than literacy alone. From a governance perspective, the application of the AIRE framework is better aligned with knowledge management (KM) systems than with traditional educational models. In institutional and policy contexts, employees and decision-makers typically engage in short-term, targeted training rather than comprehensive academic progression. Therefore, the higher levels of the taxonomy can be operationalized through KM practices such as continuous professional development, organizational learning systems, policy training modules, and decision-support frameworks. In this sense, the AIRE taxonomy serves as a bridging framework that connects educational foundations with knowledge management practices. It enables a structured transition from individual learning (micro level) to institutional capability (meso level) and governance implementation (macro level), ensuring that AI-related competencies evolve in alignment with both educational and organizational realities.

### 4.2 Socio-Technical Systems and Governance Perspective

AI systems function within intricate socio-technical networks shaped by people, institutions, and normative frameworks. From a socio-technical systems perspective, AI literacy constitutes a form of governance capacity—the ability to navigate and manage these interdependencies by understanding not only how models operate but also how they influence policy, labour, equity, and resource allocation. In this view, literacy extends beyond individual competence to become a civic infrastructure: an informed public can interpret algorithmic outcomes, challenge opaque decision-making, and anticipate systemic risks. Such literacy underpins anticipatory governance, where oversight arises through understanding rather than reaction.

Citizens, educators, and policymakers who comprehend model behaviour are therefore better positioned to mitigate ethical, social, and environmental harm. This orientation aligns with SDG 16 (Peace, Justice and Strong Institutions) and SDG 17 (Partnerships for the Goals), underscoring that knowledge and accountability are essential to sustainable governance. Building on the learning continuum outlined in the AIRE Taxonomy, these competencies evolve from individual awareness to institutional foresight—forming a conceptual bridge elaborated in Section 6.1, where literacy serves as the foundation for policy formulation and regulatory anticipation. The socio-technical system map in **Figure 8** illustrates how AI literacy operates across interconnected networks of actors, institutions, and governance norms.

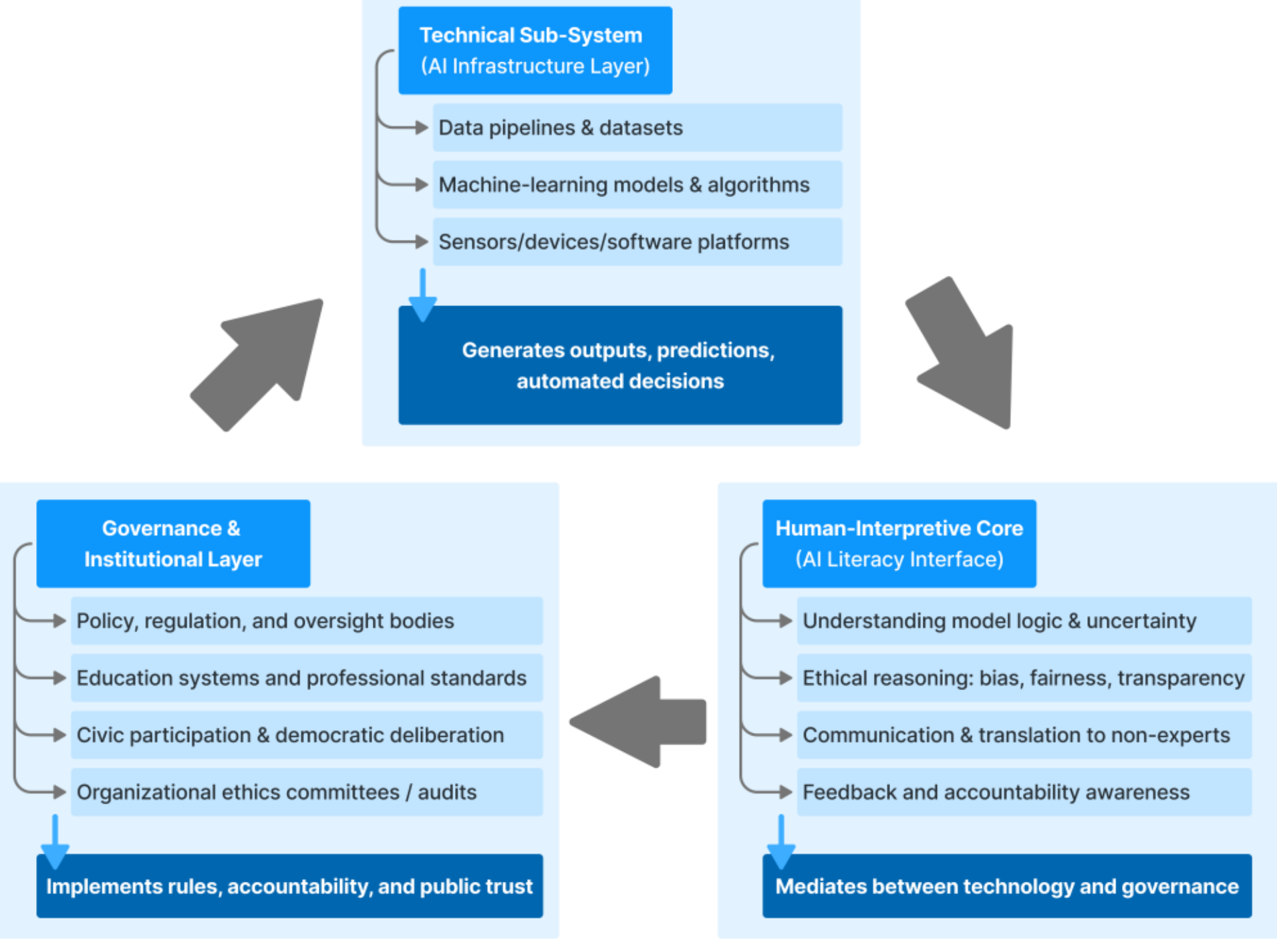

**Figure 8.** Socio-Technical AI Governance System Map

### 4.3 Macro–Meso–Micro Governance Scaling

AI literacy functions as a multilevel system of governance capacities. As shown in **Figure 9**, the AI literacy governance ecosystem flows across micro (individual), meso (institutional), and macro (policy) levels, revealing the vertical feedback that sustains ethical oversight.

- Micro level – Individual competence: students, educators, and professionals develop cognitive, ethical, and sustainability awareness.
- Meso level – Institutional practice: universities and organizations embed literacy through curricula, accreditation standards, and ethical oversight.
- Macro level – National / Global governance: governments and international bodies translate literacy into regulation, accountability frameworks, and SDG monitoring mechanisms.

These nested levels operate as feedback loops policy shapes education, education informs governance, and governance sustains ethical innovation.

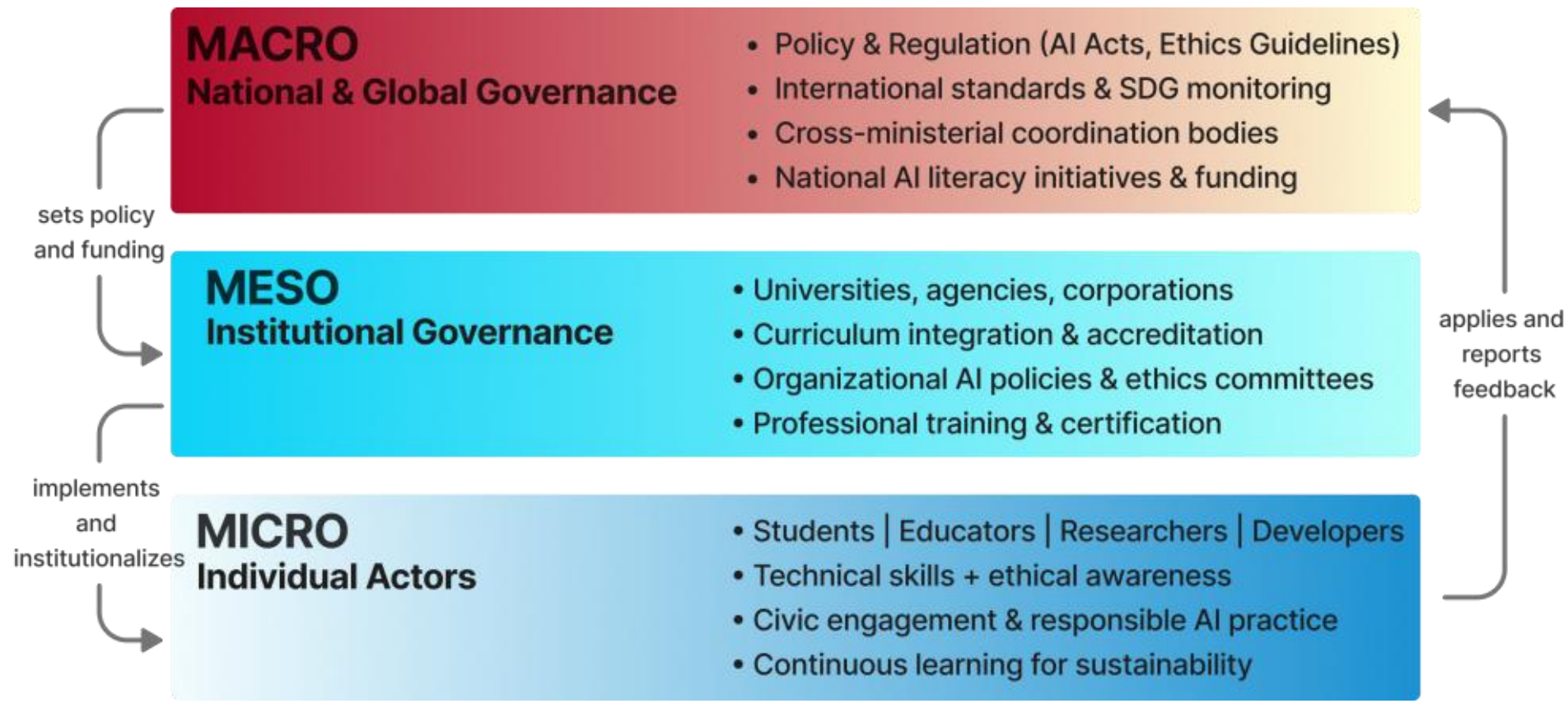

**Figure 9.** AI Literacy Governance Ecosystem Flow illustrating vertical alignment and feedback between individual (micro), institutional (meso), and national/global (macro) levels that sustain ethical and sustainable AI governance.

**Table 2** classifies AI literacy into four domains Technical & Cognitive, Ethical & Reflective, Socio-Civic & Governance, and Sustainability & Risk with sub-competencies, cognitive foci, and SDG linkages. **Figure 10** illustrates the hierarchical taxonomy of AI literacy competencies, connecting technical, ethical, civic, and sustainability literacies under one integrated framework.

**Table 2.** AI Literacy Competency Taxonomy

| Domain | Sub-Domains / Skills | Cognitive Focus | Governance Relevance | Example SDG Linkages |
|---|---|---|---|---|
| **Technical & Cognitive Literacy** | Algorithmic logic, data literacy, ML fundamentals, interpretability, automation awareness | Understanding how AI operates and produces outcomes | Supports informed policy, enables responsible system use | SDG 4 (Education), SDG 9 (Innovation) |
| **Ethical & Reflective Literacy** | Bias detection, fairness analysis, transparency, privacy, explainability, accountability | Recognizing and reasoning about moral and social implications | Shapes ethical standards and compliance cultures | SDG 5 (Gender Equality), SDG 10 (Reduced Inequalities), SDG 16 (Justice) |
| **Socio-Civic & Governance Literacy** | Legal awareness, civic participation, digital rights, misinformation resilience, democratic oversight | Enabling participation in AI-related governance and advocacy | Strengthens institutional trust and public engagement | SDG 8 (Work), SDG 11 (Cities), SDG 17 (Partnerships) |
| **Sustainability & Risk Literacy** | Environmental impacts, energy efficiency, responsible innovation, resilience planning | Anticipating long-term ecological and ethical risks | Embeds sustainability in AI governance agendas | SDG 12 (Consumption), SDG 13 (Climate), SDG 15 (Life on Land) |

This taxonomy positions AI literacy as a scaffold of ethical intelligence. The first domain establishes comprehension; the second cultivates judgment; the third extends that judgment to civic participation; and the fourth integrates global responsibility. Together, they define the capabilities needed to convert technical knowledge into ethical governance and sustainable decision-making.

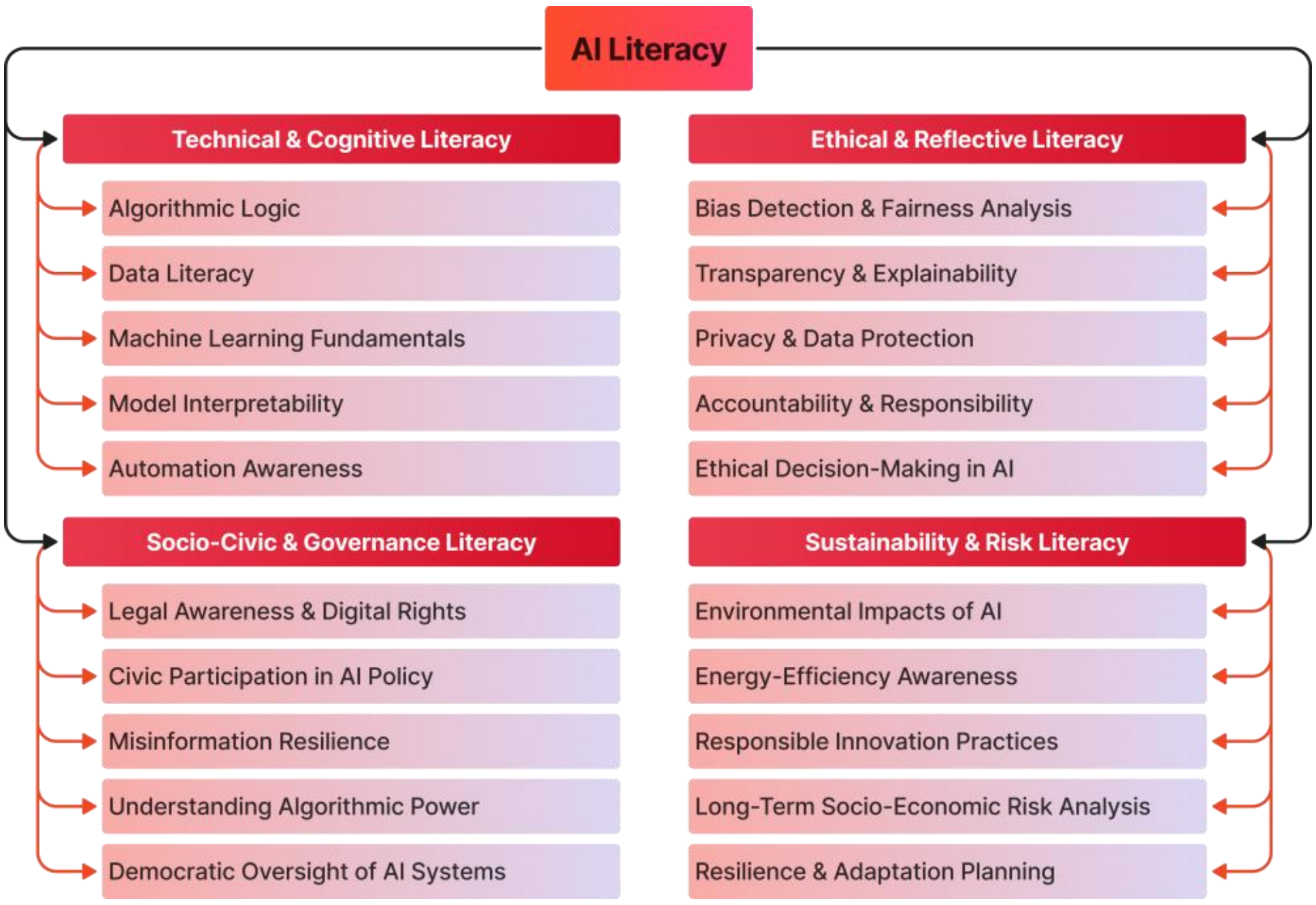

**Figure 10.** Hierarchical Taxonomy of AI Literacy Competencies

Governance levers are institutional and policy mechanisms that embed AI literacy into real-world practice. **Table 3** summarizes these levers across five governance tiers, linking them with their enabling environments and SDG implications. The hierarchical taxonomy of governance levers presented in **Figure 11** explains how policy, institutional regulation, education, technology, and community engagement jointly operationalize AI literacy.

**Table 3.** Governance Lever Taxonomy for AI Literacy Integration

| Governance Tier | Lever Type / Mechanism | Primary Function | Enabling Environment | Related SDG Targets |
|---|---|---|---|---|
| **Policy & Legislation** | National AI strategy, digital-literacy acts, data-ethics laws | Establish formal recognition of AI literacy as public competency | Government ministries, international bodies | SDG 16.6, 16.7, 17.14 |
| **Institutional Regulation** | Accreditation standards, audit requirements, algorithmic registers | Ensure literacy and ethics compliance in organizations | Universities, regulators, industry associations | SDG 4.7, 9.5 |
| **Educational Implementation** | Curriculum frameworks, teacher training, lifelong learning modules | Integrate literacy in formal and informal education | Ministries of Education, NGOs, schools | SDG 4.4, 4.7 |
| **Technological Infrastructure** | Open-data repositories, transparency dashboards, AI sandboxes | Provide accessible learning and experimentation environments | Tech agencies, R&D labs, civic-tech groups | SDG 9.C, 17.6 |
| **Community & Public Engagement** | Citizen assemblies, hackathons, awareness campaigns, participatory audits | Democratize AI governance through inclusive participation | Civil society, media, local governments | SDG 11.3, 16.10 |

This governance taxonomy operationalizes AI literacy across institutional layers. Policy creates legitimacy, education builds capacity, technology ensures transparency, and community engagement sustains accountability. When applied collectively, these levers transform literacy from a pedagogical concern into a governance instrument supporting equitable and sustainable development.

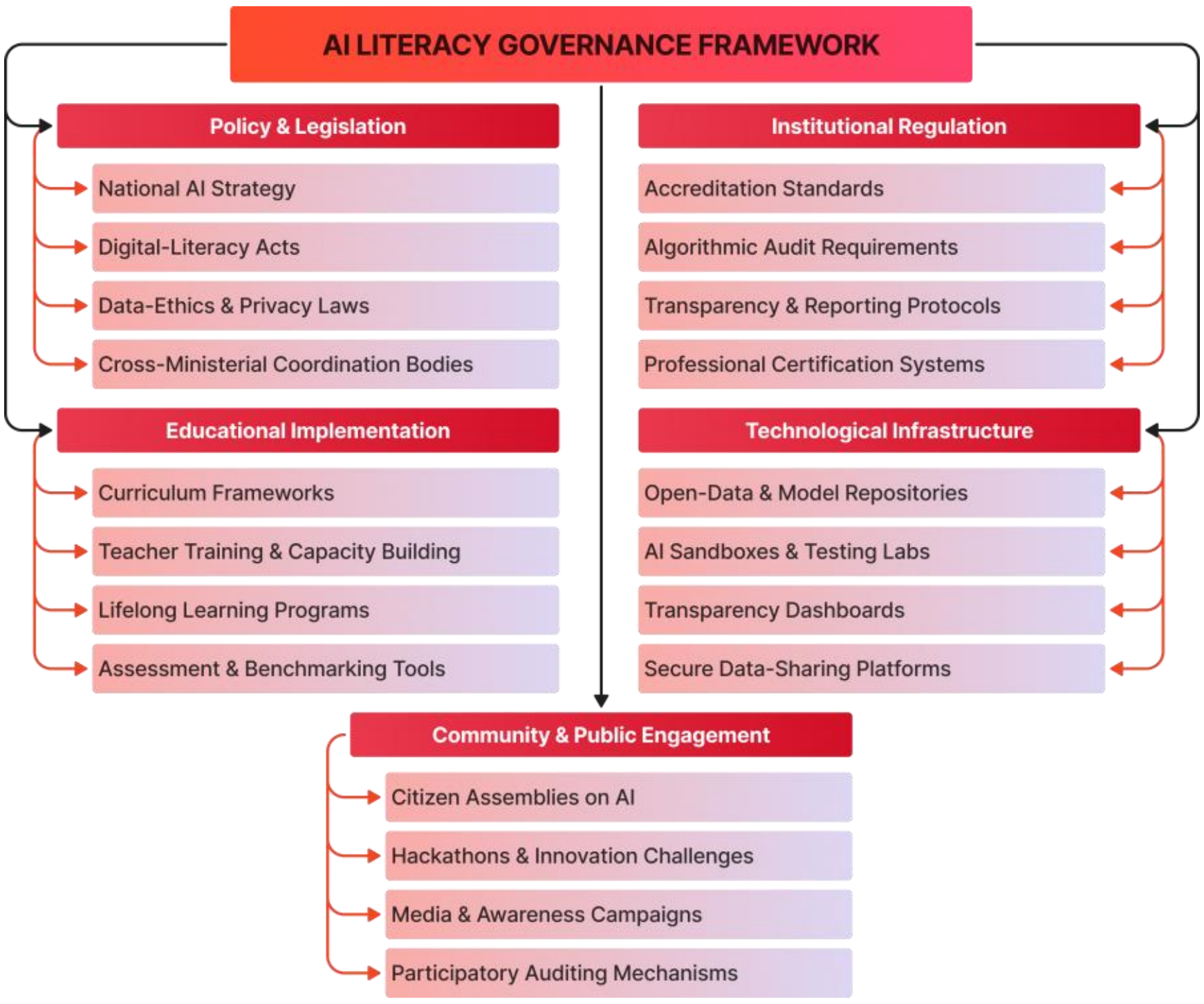

**Figure 11.** Hierarchical Taxonomy of Governance Levers

Together, these layers demonstrate the continuum from learning to governing how personal comprehension matures into institutional accountability. The following section builds on this continuum, examining how the AIRE-based competencies scale across micro (individual), meso (institutional), and macro (policy) levels within AI governance systems.

# 5. The AI–SDG Nexus and Mapping Analysis

This section operationalizes the theoretical framework through the AI–SDG Nexus, which systematically maps AI literacy domains onto all 17 Sustainable Development Goals. The mapping elucidates the intersections between competencies and governance mechanisms that collectively enable measurable sustainability outcomes.

### 5.1. Conceptual Dimensions of the Nexus

The relationship between AI literacy and the Sustainable Development Goals (SDGs) constitutes a dynamic nexus where knowledge, ethics, and governance converge to advance sustainable development. AI literacy serves as the cognitive infrastructure enabling societies to harness intelligent systems responsibly, transforming technological potential into socially beneficial outcomes. Each SDG embodies not only a thematic priority—such as education, health, or climate—but also a data-driven challenge that relies on informed human judgment. In this context, AI literacy functions as an enabling competency through which all other goals can be pursued more effectively and equitably.

This nexus is inherently multidirectional. Enhanced AI literacy strengthens the capacity of individuals and institutions to interpret algorithmic decisions, demand transparency, and design equitable AI solutions. Conversely, progress toward the SDGs raises environments conducive to cultivating such literacy; for instance, inclusive education (SDG 4) and reduced inequalities (SDG 10) expand access to AI learning opportunities, while strong institutions (SDG 16) uphold ethical governance frameworks. The interdependence between literacy and sustainability thus creates a reinforcing feedback loop that advances both technological competence and social equity. To operationalize the nexus, this study identifies five analytical dimensions linking AI literacy to the SDG agenda:

1. **Competency Dimension – Knowledge and Skills:** Technical and cognitive capacities to understand AI logic, data structures, and automation limits.
2. **Ethical Dimension – Values and Awareness:** Ability to recognize, evaluate, and mitigate bias, ensuring fairness and accountability.
3. **Governance Dimension – Policy and Oversight:** Institutional mechanisms laws, audits, procurement standards that translate literacy into enforceable ethics.
4. **Innovation Dimension – Application and Co-Creation:** Literate engagement in developing or adapting AI tools for sector-specific SDG progress.
5. **Measurement Dimension – Indicators and Evaluation:** Metrics that assess how literacy contributes to measurable SDG outcomes such as bias reduction, access equity, transparency scores.

### 5.3. Mapping Method

To operationalize the AI–SDG Nexus, the AI Literacy × SDG Matrix (**Table 4**) integrates the six tiers of the AIRE Taxonomy with four literacy domains: Technical & Cognitive, Ethical & Reflective, Socio-Civic & Governance, and Sustainability & Risk across the 17 Sustainable Development Goals. In contrast to earlier qualitative frameworks, this version introduces quantifiable indicators derived from the AIRE Taxonomy (Section 4.1), enabling systematic assessment of progress from classroom-level learning to governance-level maturity. As illustrated in **Figure 12**, the corresponding heatmap depicts the varying degrees of alignment between AI literacy competencies and individual SDGs.

### 5.4. Operational Mapping: AI Literacy × SDG Matrix

The AI Literacy × SDG mapping presented in **Table 4** is derived through a structured synthesis of literature on AI applications and governance mechanisms across SDG domains. Given the limited availability of studies explicitly using the term “AI literacy,” the mapping draws on evidence from related domains, including AI applications (climate modelling, healthcare analytics, smart infrastructure), ethical AI frameworks, and governance studies. Each SDG linkage was identified based on three criteria: (i) relevance of AI-driven decision-making, (ii) presence of data-intensive or algorithmic systems, and (iii) documented governance or ethical implications in the literature. Representative studies are included within the table to enhance transparency and traceability.

**Table 4.** Mapping of AI Literacy Competencies, Governance Levers, and AIRE-Aligned Indicators to the 17 SDGs

| SDG | Goal Focus | Key AI Literacy Domains / AIRE Tiers | Governance Levers | Expected Outcomes / Pathways | Illustrative Indicators (Policy / Education) | Representative Literature |
|---|---|---|---|---|---|---|
| **1 – No Poverty** | Digital inclusion and fair access | Technical (Recognize → Analyze) | Bias audits; digital-ID ethics | Equitable access to credit and welfare data | % AI-based targeting models audited for bias; literacy score gains in financial AI modules | (Khakurel et al. 2018) |
| **2 – Zero Hunger** | Smart agriculture & food security | Technical + Sustainability (Apply → Integrate) | Open agri-data policy; yield forecast governance | Transparent and accurate agricultural AI models | MAE of yield forecasts; % farmers trained in GeoAI literacy | (Jarzebski et al., 2023) |
| **3 – Good Health & Well-Being** | Health AI ethics and privacy | Ethical + Technical (Comprehend → Analyze) | Clinical AI approval pathways; data-ethics laws | Safer and more equitable AI-assisted care | % AI systems with bias audits; health-AI literacy certification rate | (Asim, 2025) |
| **4 – Quality Education** | Inclusive AI learning ecosystems | All domains (Recognize → Govern) | Curriculum integration; teacher training | Equity-oriented AI education programs | % schools with AIRE-based modules; teacher training completion rate | (Artyukhov et al., 2024) |
| **5 – Gender Equality** | Fairness and representation | Ethical (Analyze → Integrate) | Gender impact assessments; procurement clauses | Reduction of algorithmic gender bias | Fairness gap index; representation in dataset standards | (Jobin et al. 2019) |
| **6 – Clean Water & Sanitation** | AI for WASH and resource ethics | Sustainability (Apply → Analyze) | Water-data governance dashboards | Early contamination detection and accountability | Detection lead time improvement; public dashboard adoption rate | (Bernardo et al. 2024) |
| **7 – Affordable & Clean Energy** | AI for energy efficiency | Technical + Sustainability (Integrate) | Grid-AI audit standards | Optimized energy distribution | % energy systems audited for efficiency; energy-AI literacy score | (Khakurel et al. 2018) |
| **8 – Decent Work & Economic Growth** | Ethical automation and reskilling | Governance (Apply → Integrate) | Workplace AI impact assessments | Fair AI in employment and training | % companies with AI ethics policies; reskilling program coverage | (Dwivedi et al. 2023) |
| **9 – Industry, Innovation & Infrastructure** | Responsible MLOps and safety | Technical + Governance (Analyze → Govern) | Algorithmic risk tiers; incident reporting | Safe and transparent AI innovation | % firms filing AI risk disclosures; incident response time | (Arora & Mishra 2019) |
| **10 – Reduced Inequalities** | Accessibility & inclusive AI | Ethical + Civic (Apply → Integrate) | Equality impact assessments | Fair and multilingual AI services | Accessibility certification rate; bias-reduction score | (Pendyala 2024) |
| **11 – Sustainable Cities & Communities** | Urban AI ethics and mobility | Civic + Governance (Integrate → Govern) | Smart-city AI charters; public participation | Transparent urban AI systems | % deployments with public audits; citizen participation index | (Al-Raeei 2024) |
| **12 – Responsible Consumption & Production** | Circular economy & traceability | Sustainability (Analyze → Integrate) | ESG AI traceability standards | Transparent supply chains and waste reduction | % suppliers with traceable AI logs; greenwashing detection precision | (Khakurel et al. 2018) |
| **13 – Climate Action** | Climate model literacy & adaptation | Sustainability + Ethical (Integrate → Govern) | Open climate-AI benchmarks | Trust in climate predictions | Forecast transparency index; adaptation policy integration rate | (Filho et al. 2025) |
| **14 – Life Below Water** | Marine AI governance | Technical + Sustainability (Apply → Integrate) | Blue-data policy; bycatch monitoring rules | Reduced illegal fishing and marine loss | % monitored zones; biodiversity AI accuracy score | (Jarzebski et al. 2023) |

| 15 – Life on Land | Conservation AI ethics | Sustainability (Analyze → Govern) | Citizen-science data rights | Biodiversity protection and trust | % protected areas with AI monitoring; incident response speed | (Jarzebski et al. 2023) |
|---|---|---|---|---|---|---|
| **16 – Peace, Justice & Strong Institutions** | Algorithmic accountability and trust | Governance (Integrate → Govern) | Public audit portals; AI accountability laws | Transparent decision-making systems | % AI systems with audit logs; public report disclosure rate | (Floridi & Cowls 2019) |
| **17 – Partnerships for the Goals** | Data collaboration & shared metrics | Governance + Ethical (Integrate → Govern) | Data trusts and interoperability standards | Ethical cross-sector collaboration | # of active data trusts; SDG indicator coverage index | (OECD 2024; UNESCO 2023) |

The inclusion of representative literature strengthens the interpretability of the mapping and demonstrates alignment with existing research across AI application domains.

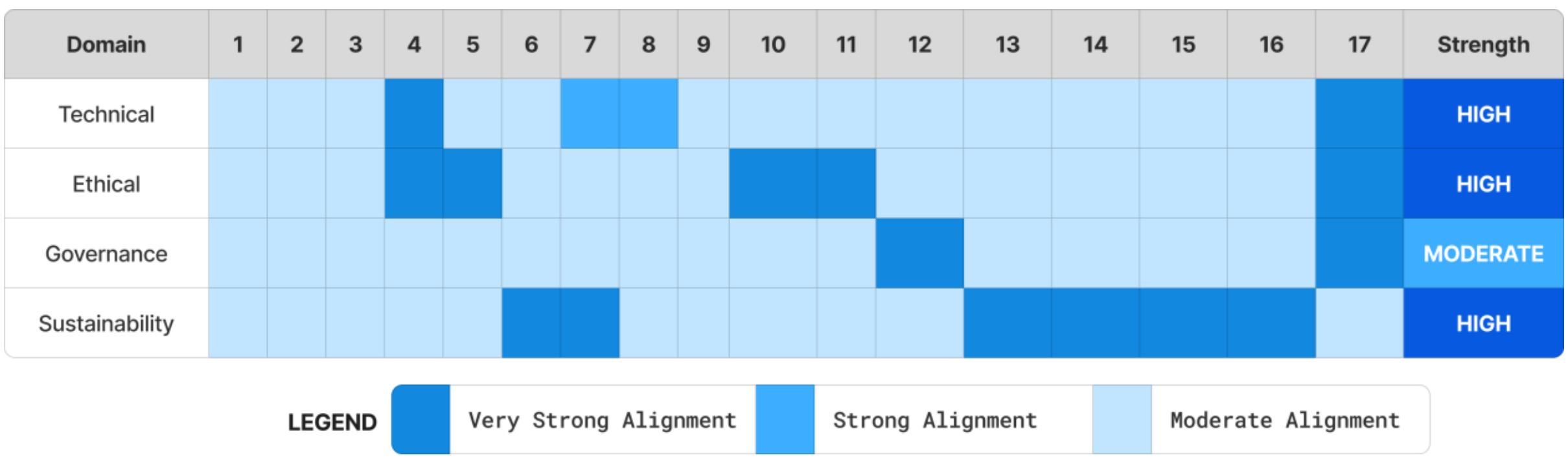

**Figure 12.** Heatmap of AI Literacy Domain Alignment with the 17 Sustainable Development Goal

### 5.5. Interpretation of the Mapping

**Table 5** mapping interpretation reveals that AI literacy contributes to every SDG through distinct yet complementary pathways. Across all clusters, AI literacy operates as a governance multiplier—linking individual competence with institutional accountability. It translates ethical and reflective understanding into measurable sustainability performance, illustrating how literacy evolves from pedagogical development to policy integration within the AI–SDG Nexus.

**Table 5.** Interpretation of Mapping of AI Literacy Contributions Across Sustainable Development Goal (SDG) Clusters

| SDG Cluster | Key SDGs | AI Literacy Dimensions and Competencies | Governance and Societal Implications |
|---|---|---|---|
| **Foundational Enablers** | SDGs 1–5 | Data ethics, bias detection, algorithmic fairness, financial AI awareness, equitable dataset design | Promotes poverty reduction, inclusive education, gender equality, and transparent service delivery; empowers marginalized populations through fair data practices. |
| **Infrastructure and Innovation Goals** | SDGs 6–9 | Technical literacy in IoT, machine learning, and optimization; governance tools such as open data standards and algorithmic audits | Enhances efficiency and accountability in water, energy, and industrial systems; supports transparent and ethical technological innovation. |
| **Equity and Inclusion Goals** | SDGs 10–11 | Accessibility-focused AI literacy, localization and cultural adaptation skills | Fosters equitable access to urban mobility, housing, and information; advances inclusive, culturally responsive cities and digital ecosystems. |
| **Sustainability and Environmental Goals** | SDGs 12–15 | Environmental data literacy, life-cycle assessment, remote sensing, and uncertainty interpretation | Enables responsible production, biodiversity protection, and climate adaptation; builds public trust in environmental and climate models. |
| **Governance and Partnership Goals** | SDGs 16–17 | Civic-tech literacy, algorithmic accountability, data-collaboration and ethical data-sharing skills | Strengthens institutions, fosters transparent governance, and promotes cross-sector partnerships for global monitoring and innovation. |

The mapping illustrates an integrated network of relationships spanning all seventeen SDGs. Ethical reasoning emerges as a unifying dimension, underscoring that technical proficiency alone is insufficient for sustaining responsible innovation. Equally critical is governance coherence without transparent policies, auditing mechanisms, and open-data standards, AI literacy remains an individual competency rather than a collective social capacity. Measurement serves as the linkage between learning and accountability, where indicators such as bias-reduction rates, accessibility compliance, and traceability scores transform literacy into measurable progress. Education functions as the central catalyst uniting these dimensions; when curricula integrate ethics and sustainability with computational proficiency, the classroom becomes the foundational layer of governance. The reciprocal relationship between literacy and development further reinforces this dynamic: fostering literacy promotes ethical innovation, while sustainable progress deepens societal comprehension of AI. Collectively, these findings position AI literacy not as a discrete educational initiative but as an evolving governance mechanism through which the SDGs gain ethical, technical, and civic coherence.

### 5.6. Implications for Framework Development

The AI-SDG nexus underscores that sustainable progress relies on integrating AI literacy into governance ecosystems. This mapping enables policymakers to identify strategic leverage points such as national curricula, professional training programs, and regulatory frameworks that translate literacy into measurable SDG outcomes. For educators, the matrix serves as a curriculum blueprint aligning technical education with global sustainability priorities, while for institutions, it functions as a self-assessment tool to evaluate ethical capacity and compliance preparedness. By framing AI literacy as both a governance competency and a driver of sustainability, the framework reconceptualizes education as policy infrastructure vital to fostering ethical, resilient, and human-cantered technological advancement.

# 6. Governance-Oriented Framework and Policy Implications

The mapping between AI literacy and the Sustainable Development Goals (SDGs) reveals a deeper truth: the sustainability of technological societies depends less on what AI can do and more on how people understand and govern it. Governance is not an afterthought to innovation but its living architecture the structure through which intelligence, human and artificial, aligns with shared ethical and developmental goals. In this context, AI literacy becomes both the language and the lens of governance, translating complex technical processes into accountable, transparent, and sustainable practice.

### 6.1. Reframing Governance through AI Literacy

Traditional governance frameworks typically respond to risks only after harm has occurred. A literacy-driven approach inverts this paradigm, positioning foresight as the core of regulation. When citizens, educators, and policymakers share a foundational understanding of algorithmic reasoning, they can anticipate vulnerabilities, demand transparency, and collaboratively design preventive mechanisms. The AIRE Taxonomy embodies this transformation: the individual progression from Recognize to Govern parallels the institutional shift from awareness to anticipatory oversight. In such literate societies, regulation emerges not solely through legal instruments but through collective comprehension—making learning the first act of governance. The sustainability of technological societies therefore depends less on the capabilities of AI than on the depth of human understanding that guides its use. Governance becomes not a static structure applied post-innovation, but a dynamic architecture in which human and artificial intelligence co-evolve within an ethical framework. Through this lens, AI literacy functions as both the language and the lens of modern governance, translating complex systems into transparent, accountable, and sustainable practice.

### 6.2. Policy Integration Pathways

Embedding AI literacy within policy frameworks requires coordination across education, regulation, and innovation systems. National AI strategies should treat literacy as a public good, equivalent in significance to health or environmental education. Policy integration can occur through three mutually reinforcing routes:

1. **Education Systems** – Integrate AIRE-based literacy outcomes into national curricula and professional training; fund teacher preparation and open learning resources.
2. **Regulatory Design** – Align data-protection, transparency, and algorithmic-audit requirements with literacy indicators that measure institutional competence, not only technical compliance.
3. **Institutional Accreditation** – Include ethical-literacy benchmarks such as transparency reporting, bias audits, and explainability protocols in accreditation and quality-assurance standards.

Through these routes, education and policy co-evolve: policy legitimizes literacy, and literacy empowers policy.

### 6.3. Inclusivity and Public Engagement

Without inclusive access, AI literacy risks reinforcing the very divides it seeks to close. Programs must therefore prioritize gender equity, linguistic diversity, and accessibility for marginalized communities. Partnerships among universities, civil society, and industry can create community innovation labs participatory spaces where citizens learn to question and co-design the AI systems shaping their lives. Public engagement also strengthens legitimacy. Literate citizens are better equipped to contribute to policy debates, evaluate media narratives, and participate in algorithmic audits. When literacy expands democratically, governance becomes a shared civic practice rather than a technocratic imposition.

### 6.4. Building an Accountability Ecosystem

A literacy-based governance model reframes accountability as mutual understanding rather than top-down enforcement.

- **Developers** must design for transparency and explainability.
- **Educators** must teach interpretive and ethical reasoning.
- **Policymakers** must ensure that oversight mechanisms remain understandable to the public.

When these roles align, governance becomes an ecosystem of shared responsibility instead of segmented compliance. AI literacy thus operates as the connective tissue linking the technical, the ethical, and the civic into one sustainable system.

### 6.5. Curriculum Feasibility and Scalable Implementation

While the AIRE taxonomy outlines a comprehensive progression of AI-related competencies, its implementation within formal education systems must account for curriculum constraints and student workload. In long-duration educational settings such as K–12 or undergraduate programs, introducing additional standalone content on governance, sustainability, and AI risks overburdening learners. To address this challenge, the framework is designed to be integrative rather than additive. Instead of introducing new subjects, AI literacy components can be embedded within existing disciplines such as science, social studies, technology, and civic education. For example, basic AI awareness (Recognize, Comprehend) can be incorporated into digital literacy modules, while ethical and societal implications can be discussed within existing civic or social science curricula. Furthermore, the progression defined by the AIRE taxonomy should be implemented in a layered and age-appropriate manner. Foundational competencies corresponding to AI literacy can be introduced in early and middle education, while higher-order competencies such as critical evaluation and governance awareness are gradually developed at advanced stages of education or through professional training. This approach ensures that students are not required to master all dimensions simultaneously but instead build competencies progressively over time. It also aligns with knowledge management principles, where learning is distributed across stages, contexts, and roles rather than concentrated within a single curriculum. Therefore, the AIRE framework should be interpreted as a scalable and flexible guideline that supports curriculum integration, lifelong learning, and institutional adaptation, rather than a fixed or mandatory instructional sequence.

### 6.6. Global and Long-Term Implications

The proposed governance framework emphasizes that literacy should be regarded not as a transient training initiative but as a strategic capacity essential for sustainable development. Integrating literacy indicators into SDG monitoring mechanisms would enable nations to evaluate not only the technologies they adopt but also the effectiveness and responsibility with which these technologies are governed. Collaborative efforts among international bodies such as UNESCO, the OECD, and the UN AI–SDG Lab could facilitate the development of standardized benchmarks for assessing AI literacy readiness within broader governance maturity indices.

To validate the conceptual model's practical relevance, an empirical assessment was conducted through a cross-sector survey involving 300 participants. The survey examined AI literacy readiness and its perceived relationship to SDG implementation. The findings, presented in the subsequent section, evaluate the extent to which the proposed governance-oriented framework reflects real-world awareness and institutional capacity.

# 7. Empirical Validation of AI Literacy Readiness and SDG Integration

To empirically substantiate the conceptual AI–SDG Nexus, a structured cross-sectional survey was administered to 300 respondents. The statistical analysis followed a three-stage approach: descriptive analysis to assess literacy readiness, correlation analysis to examine inter-domain relationships, and regression modelling to evaluate predictors of nexus awareness. This integrated approach ensures consistency between the methodological framework and empirical findings.

### 7.1. Survey Design and Participant Profile

A cross-sector survey (n = 300) was conducted within a single-country context (Bangladesh), capturing perspectives from students (40%), educators (25%), professionals (20%), and policymakers (15%). Participants had moderate-to-high digital exposure (64 % daily AI use; 42 % formal training). Bangladesh represents a relevant case for examining AI literacy in emerging economies, where rapid digital transformation coexists with evolving governance capacity. The goal was to assess domain-wise literacy readiness and perceived SDG linkages using the five-dimension AIRE-based framework. As shown in **Figure 13** summarizes the participant composition by professional background, confirming balanced representation among students, educators, professionals, and policymakers. All survey procedures adhered to standard ethical research practices. Participation was voluntary and based on informed consent, with respondents made aware of the study's purpose and use of data for academic research. No personally identifiable information was collected, and all responses were anonymized and analyzed in aggregate form to ensure confidentiality and data protection. While policymakers constitute 15% of the sample, the study adopts a cross-sector perspective to capture AI literacy as a distributed competency across societal roles, including students, educators, professionals, and policymakers. This approach reflects the conceptualization of AI literacy as a multi-level governance capacity, where readiness is not confined to policymakers alone but emerges from interactions across micro, meso, and macro levels. Therefore, the empirical analysis aims to provide indicative insights into cross-domain readiness rather than a policy-exclusive evaluation.

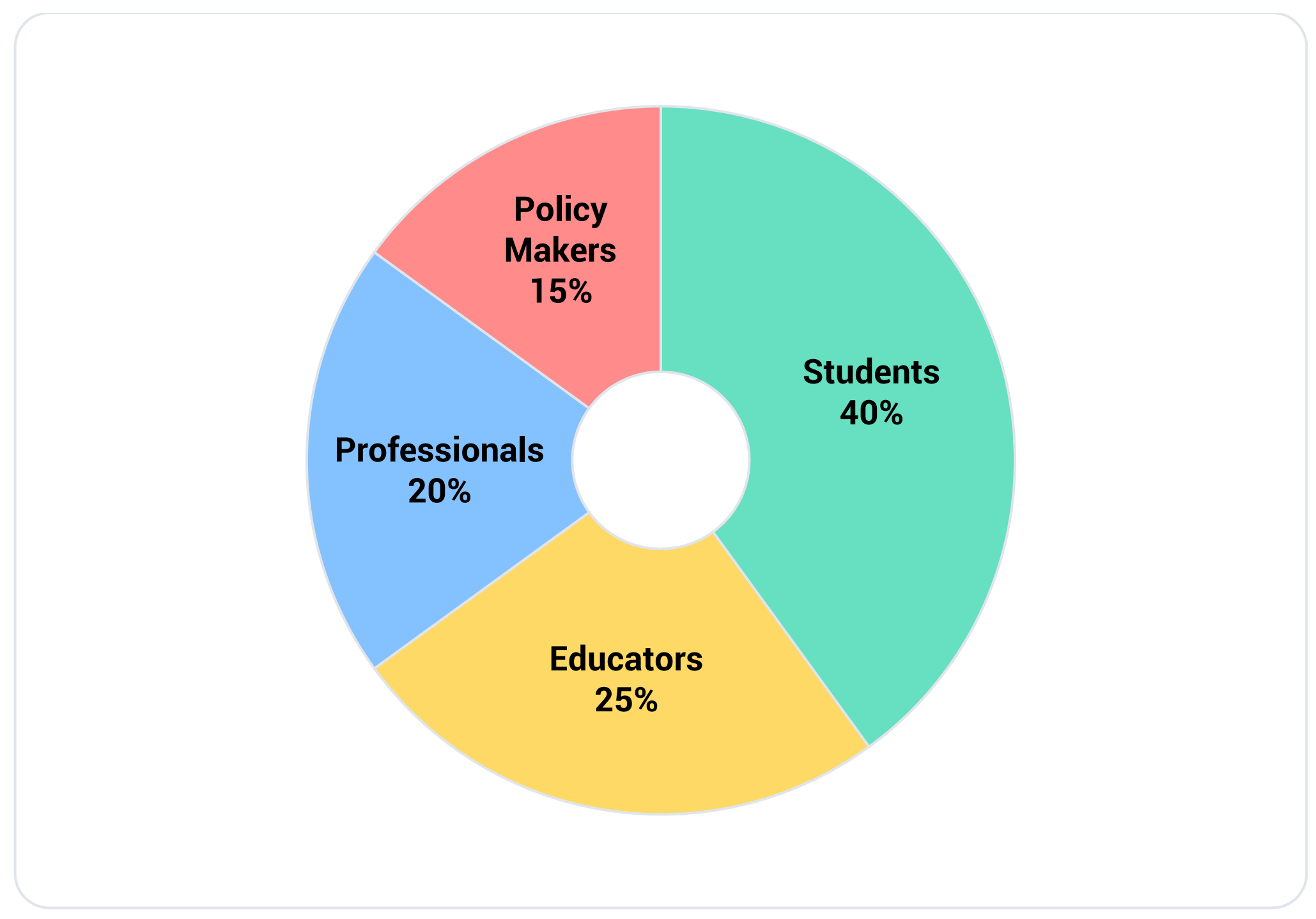

**Figure 13.** Participant Composition by Professional Background (n = 300)

### 7.2. Domain-Wise Literacy Readiness

The composite AI–SDG Literacy Index (ASLI) averaged 3.59/5, indicating moderate readiness. The AI–SDG Literacy Index (ASLI) represents the overall literacy score derived from four domains: Technical and Cognitive Literacy (TL), Ethical and Reflective Literacy (EL), Governance and Civic Literacy (GL), and Sustainability and Risk Literacy (SL). Each domain score was calculated as the mean of its corresponding Likert-scale items (1–5), and the ASLI was computed as the average of these four domain scores. The Nexus Awareness Index (NAI) captures respondents' perceived understanding of the relationship between AI literacy and the Sustainable Development Goals (SDGs). It was computed as the mean score of the items in Section F of the survey instrument. Higher values indicate stronger awareness of AI–SDG interconnections. All indices were normalized on a 1–5 scale for interpretability and comparability across domains. The aggregation of items into domain-level indices is justified by their strong internal consistency and conceptual alignment. A preliminary factor inspection indicated that items cluster within their intended domains (technical, ethical, governance, sustainability, and nexus awareness), supporting the construct validity of the composite measures. The internal consistency of the survey instrument was evaluated using Cronbach's alpha to ensure the reliability of the constructed indices. The results indicate high reliability across all domains, with Technical Literacy ($\alpha = 0.94$), Ethical Literacy ($\alpha = 0.93$), Governance Literacy ($\alpha = 0.94$), and Sustainability Literacy ($\alpha = 0.94$) demonstrating excellent internal consistency. The Nexus Awareness Index (NAI) also showed acceptable reliability ($\alpha = 0.76$), exceeding the recommended threshold of 0.70. These findings confirm that the items within each domain consistently measure the intended constructs, thereby supporting the aggregation of items into composite indices such as ASLI and NAI for subsequent analysis. Technical literacy ranked highest (M = 3.87), followed by sustainability (3.55) and ethics (3.42); governance literacy was lowest (3.21). This pattern reflects strong technical awareness but weaker policy and accountability skills. The radar chart in **Figure 14** compares average literacy readiness across domains, revealing that technical literacy is highest while governance literacy remains weakest.

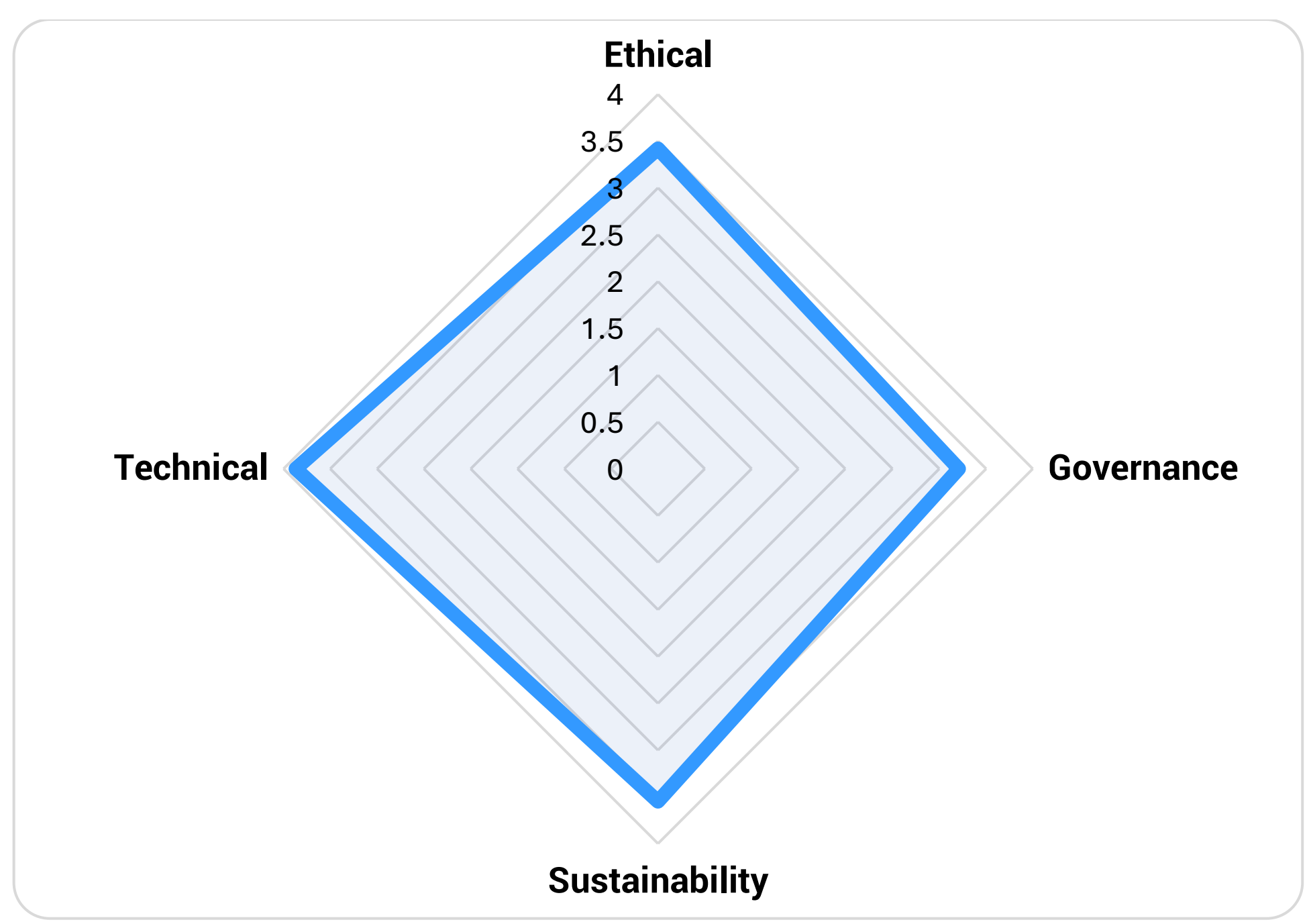

**Figure 14.** Radar Chart of AI Literacy Domains (1–5 scale, n = 300).

### 7.3. Perceived SDG Impact and Correlational Insights

From **Figure 15** it is clear that respondents prioritized actions aligned with SDG 16**,** with public awareness (24.5%) and government policy (18.8%) emerging as the strongest drivers of AI literacy. SDG 4 followed through curriculum training (21.5%), while SDG 9 gained support via industry partnerships and innovation labs (combined 22.8%). Open data, tied to SDG 17, accounted for 12.4%. As shown in **Figure 15**, governance, education, and innovation form the core leverage points for advancing AI literacy—reinforcing the AIRE framework's insight that sustainable AI adoption depends primarily on ethical and governance capacities, not technical fluency alone.

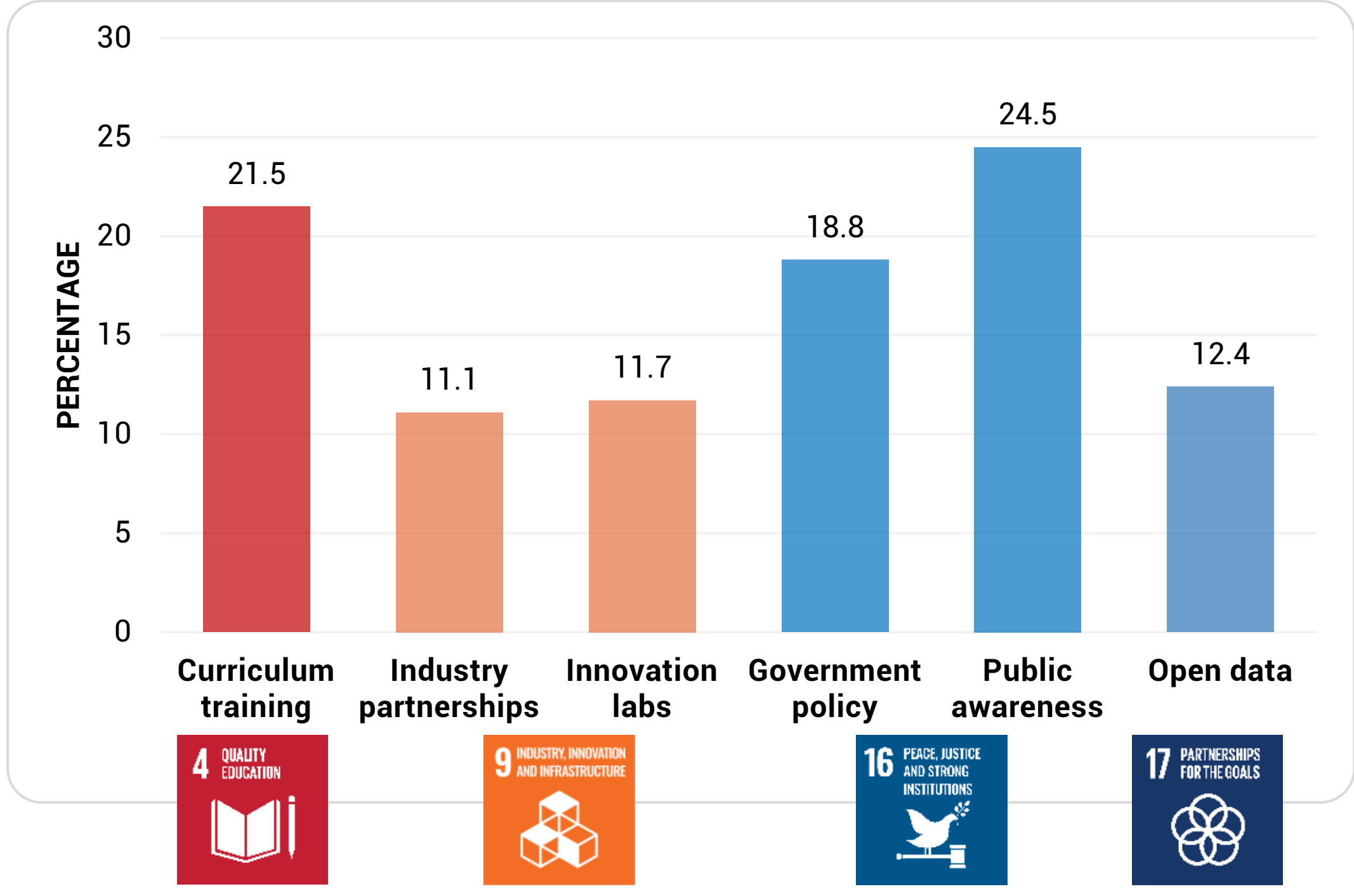

**Figure 15.** Perceived AI Impact Across Selected SDGs and Corresponding Literacy Domains.

Correlation analysis produced a consistent pattern of interdependence: Governance Literacy (GL) and Nexus Awareness (NAI) displayed the strongest association ($r = 0.67$, $p < 0.01$), followed by Ethical and Sustainability literacy ($r = 0.56$). These findings suggest that citizens who understand AI's ethical and policy dimensions are more likely to recognize its potential for sustainable development. Correlation patterns in **Figure 16** confirm the close coupling between governance competence and holistic SDG awareness. These correlational findings provide preliminary evidence of association but do not imply causal relationships, consistent with the cross-sectional design.

| Domain | Technical | Ethical | Governance | Sustainability | Nexus |
|---|---|---|---|---|---|
| Technical | 1 | 0.49 | 0.41 | 0.44 | 0.55 |
| Ethical | 0.49 | 1 | 0.62 | 0.56 | 0.6 |
| Governance | 0.41 | 0.62 | 1 | 0.58 | 0.67 |
| Sustainability | 0.44 | 0.56 | 0.58 | 1 | 0.64 |
| Nexus | 0.55 | 0.6 | 0.67 | 0.64 | 1 |

**Figure 16.** Inter-Domain Correlation Heatmap–Colour intensity denotes relationship strength between literacy dimensions.

### 7.4. Barriers and Enablers of AI Literacy Integration

The most significant constraint identified by respondents is lack of training (60.2%), followed by ethical mistrust (49.2%) and low institutional support (48.3%). Policy gaps (39.8%) and high cost (39.8%) also represent notable structural obstacles, while limited local-language content remains a minor but meaningful concern (5.9%). As shown in **Figure 17**, these patterns reveal that skill shortages and trust deficits overshadow financial or policy-related barriers, underscoring the need for stronger training ecosystems and institutional support to advance AI literacy effectively.

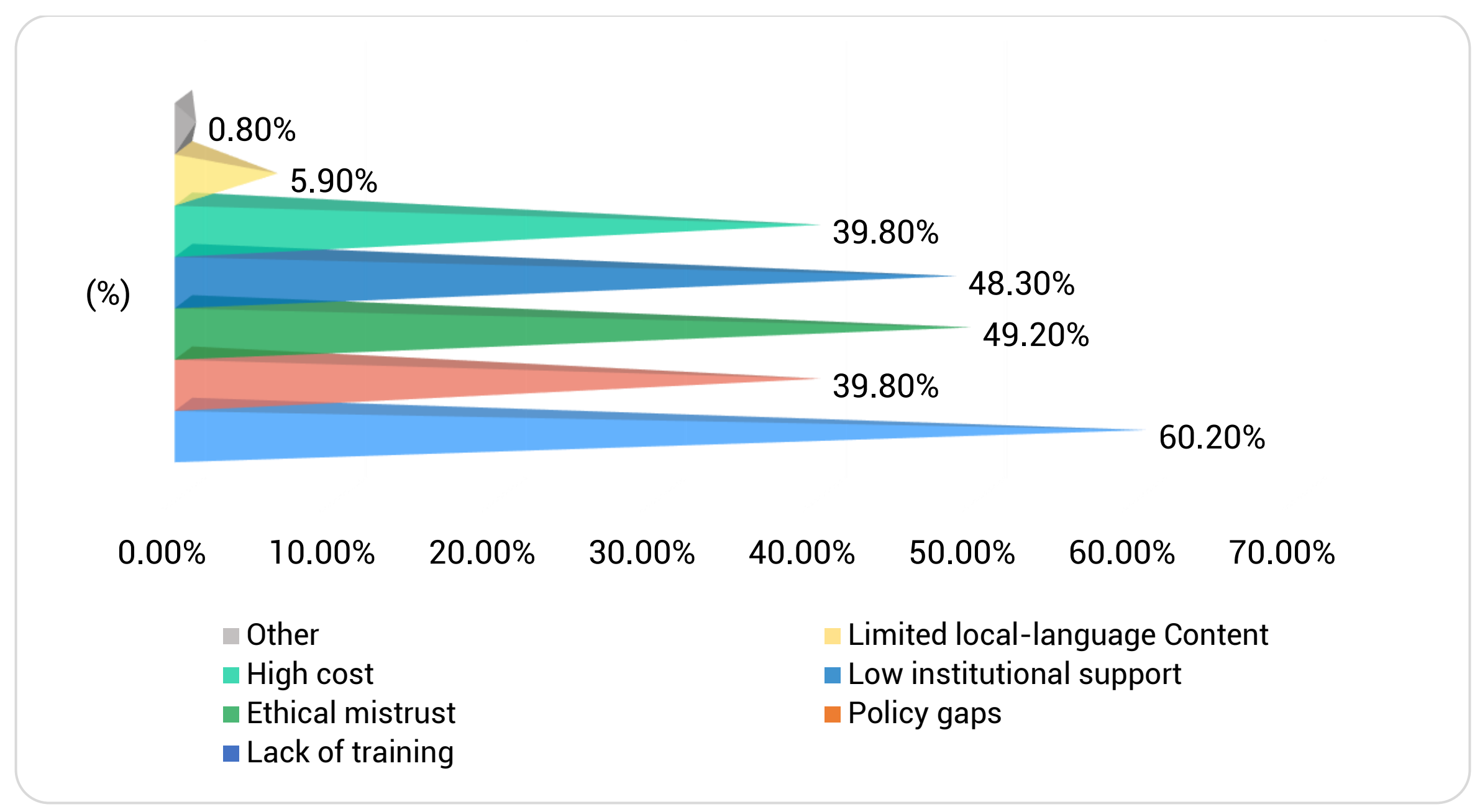

**Figure 17.** Barriers and Enablers of AI Literacy for SDG Integration chart comparing structural deficits with governance drivers.

### 7.5 Cross-Domain and Policy Interpretation

To further examine these relationships, a multiple linear regression model was estimated with Nexus Awareness Index (NAI) as the dependent variable and literacy domains as predictors. The results shown in **Table 6** indicate that governance literacy is the strongest predictor of nexus awareness ($\beta = 0.64$, $p < 0.01$), followed by ethical and technical literacy. As per **Table 7**, the model explains 43% of the variance ($R^2 = 0.43$), suggesting substantial explanatory power. Importantly, these findings should be interpreted as predictive associations rather than causal effects, given the cross-sectional nature of the data.

**Table 6.** Domain-wise Literacy Readiness by Professional Group

| Professional Group | TL (Technical) | EL (Ethical) | GL (Governance) | SL (Sustainability) | NAI (Nexus Awareness) |
|---|---|---|---|---|---|
| Students (40%) | 3.94 ± 0.71 | 3.45 ± 0.68 | 3.09 ± 0.63 | 3.48 ± 0.65 | 3.61 ± 0.66 |
| Educators (25%) | 3.81 ± 0.69 | 3.38 ± 0.64 | 3.25 ± 0.59 | 3.62 ± 0.61 | 3.70 ± 0.64 |
| Professionals (20%) | 3.79 ± 0.67 | 3.47 ± 0.71 | 3.30 ± 0.60 | 3.58 ± 0.63 | 3.73 ± 0.60 |
| Policymakers (15%) | 3.91 ± 0.66 | 3.54 ± 0.70 | 3.46 ± 0.58 | 3.68 ± 0.59 | 3.82 ± 0.57 |
| Overall Mean | **3.87** | **3.46** | **3.28** | **3.59** | **3.72** |

Governance literacy steadily increases with professional responsibility; policymakers rank highest (M = 3.46) while students remain lowest (M = 3.09). This pattern echoes the conceptual model's vertical scaling from micro (individual) to macro (policy) competencies.

**Table 7.** Predictive Model of Governance Literacy → Nexus Awareness

| Predictor | β (Standardized) | t | p | $R^2$ |
|---|---|---|---|---|
| Governance Literacy (GL) | 0.64 | 8.47 | < 0.001 | 0.43 |
| Ethical Literacy (EL) | 0.21 | 2.93 | 0.004 | – |
| Technical Literacy (TL) | 0.18 | 2.61 | 0.010 | – |

The results show a clear progression from technical understanding to governance foresight, confirming the AIRE model's continuum from learning to policy action. Governance literacy emerges as the strongest driver of SDG awareness, while policy support and infrastructure access stand out as key enablers. These insights affirm that AI literacy functions not just as knowledge, but as governance in practice—bridging education, ethics, and sustainability toward the goals of the 2030 Agenda. Therefore, the findings should be interpreted as indicative rather than globally representative, providing a foundation for future cross-country validation studies.

### 7.6 Limitations and Scope of Empirical Validation

This study includes an empirical component based on a cross-sectional survey designed to provide initial insights into AI literacy readiness across different stakeholder groups. However, several limitations should be acknowledged. First, the sample is context-specific and does not represent global variation in institutional capacity, governance maturity, or technological infrastructure. Second, policymakers constitute a relatively smaller proportion of the sample (15%), which limits the extent to which governance-specific conclusions can be generalized. Third, the cross-sectional design captures perceptions at a single point in time and does not reflect longitudinal changes in AI literacy development. Also, the study is based on a single-country sample (Bangladesh), which may limit the generalizability of empirical findings across different governance and socio-economic contexts. Therefore, the empirical findings should be interpreted as exploratory and indicative rather than definitive validation of the proposed framework. Future research should include broader geographic samples, higher representation of policymakers, and longitudinal or mixed-method approaches to strengthen empirical validation.

# Conclusion

This study examines the role of AI literacy in advancing sustainable development within the framework of the United Nations Sustainable Development Goals (SDGs), addressing the growing need to move beyond technical proficiency toward ethical and governance-oriented competencies. The findings demonstrate that AI literacy functions as a governance capacity rather than merely an educational outcome, enabling responsible and sustainable AI deployment. Through the proposed AIRE Taxonomy, the study establishes a progression from foundational awareness to strategic governance, linking individual learning with institutional accountability, while the AI–SDG Nexus Framework illustrates how AI literacy may be understood as an "18th SDG" in a heuristic sense by supporting all 17 SDGs. Empirical results from a cross-sector survey (n = 300) reveal that although technical literacy is relatively strong, ethical and governance literacy remain underdeveloped, with governance literacy emerging as the strongest predictor of SDG awareness. These findings contribute theoretically by reconceptualizing AI literacy as a multidimensional governance construct and practically by highlighting the need to embed literacy into education systems, institutional frameworks, and policy design. However, the empirical findings are based on a single-country sample, which may limit the generalizability of the results across different regional or cultural contexts. While the empirical findings offer initial support for the proposed framework, they should be interpreted as context-specific and exploratory. Broader validation across diverse geographic and institutional settings, particularly with greater representation of policymakers, is required to fully assess the framework's applicability in governance contexts. Future research should focus on longitudinal validation, cross-country analysis, and the development of standardized AI literacy indicators for integration into SDG monitoring systems, as well as the application of the AIRE framework across sector-specific governance contexts.

## Data Availability Statement

The data that support the findings of this study are not publicly available due to confidentiality and ethical considerations, as the dataset contains survey responses collected under conditions of anonymity. However, anonymized data may be made available by the authors upon reasonable request for academic and research purposes.

# Appendix A

## AI Literacy & Sustainable Development Goals (SDG) Readiness Survey

The empirical analysis in this study comprises three components:

- **Descriptive Statistics:** Mean and standard deviation values were calculated for all literacy domains (TL, EL, GL, SL) and Nexus Awareness Index (NAI) to assess overall readiness levels.
- **Correlation Analysis:** Pearson correlation coefficients were computed to examine relationships among literacy domains and their association with SDG awareness.
- **Regression Analysis:** A multiple linear regression model was conducted to evaluate the predictive influence of AI literacy domains on nexus awareness (NAI). Governance, ethical, and technical literacy were included as independent variables.

This multi-stage analytical approach ensures alignment between descriptive insights, relational patterns, and predictive modelling within the AI–SDG Nexus framework. All results are interpreted as associative due to the cross-sectional design.

# Section A: Demographic and Background Information

1. **What is your age group?** □ 18–24 □ 25–34 □ 35–44 □ 45–54 □ 55+
2. **Gender:** □ Male □ Female □ Prefer not to say
3. **Highest education level completed:**
   □ High school □ Undergraduate □ Graduate □ Ph.D. □ Other: _____
4. **Professional background:**
   □ Student □ Educator □ Researcher □ Policy □ Tech Industry □ Other: ____
5. **Have you taken any AI-related course or training?**
   □ Yes □ No
6. **How frequently do you use AI tools (e.g., ChatGPT, Gemini, others)?**
   ① Never ② Rarely ③ Sometimes ④ Often ⑤ Daily

# Section B: Technical and Cognitive AI Literacy (SDG 4, 8, 9)

**1. I understand how AI systems make predictions or recommendations.**
① Strongly Disagree ② Disagree ③ Neutral ④ Agree ⑤ Strongly Agree
**2. I can identify when an AI model might produce biased results.**
① Strongly Disagree ② Disagree ③ Neutral ④ Agree ⑤ Strongly Agree
**3. I have applied AI tools in my learning or work.**
① Strongly Disagree ② Disagree ③ Neutral ④ Agree ⑤ Strongly Agree
**4. AI can enhance quality education (SDG 4).**
① Strongly Disagree ② Disagree ③ Neutral ④ Agree ⑤ Strongly Agree
**5. AI will create innovative and sustainable industries (SDG 9).**
① Strongly Disagree ② Disagree ③ Neutral ④ Agree ⑤ Strongly Agree

# Section C: Ethical and Reflective Literacy (SDG 5, 10, 16)

**6. AI systems can reflect gender or cultural biases**.
① Strongly Disagree ② Disagree ③ Neutral ④ Agree ⑤ Strongly Agree
**7. All AI systems should be audited for fairness.**
① Strongly Disagree ② Disagree ③ Neutral ④ Agree ⑤ Strongly Agree
**8. Transparency increases public trust in AI.**

① Strongly Disagree ② Disagree ③ Neutral ④ Agree ⑤ Strongly Agree

**9. Ethical guidelines are needed to prevent misuse.**

① Strongly Disagree ② Disagree ③ Neutral ④ Agree ⑤ Strongly Agree

**10. AI literacy should include privacy and accountability training.**

① Strongly Disagree ② Disagree ③ Neutral ④ Agree ⑤ Strongly Agree

# Section D: Socio-Civic and Governance Literacy (SDG 1, 11, 16, 17)

**11. Citizens should have the right to understand AI decisions.**

① Strongly Disagree ② Disagree ③ Neutral ④ Agree ⑤ Strongly Agree

**12. I am familiar with AI governance policies.**

① Strongly Disagree ② Disagree ③ Neutral ④ Agree ⑤ Strongly Agree

**13. AI can increase transparency in governance.**

① Strongly Disagree ② Disagree ③ Neutral ④ Agree ⑤ Strongly Agree

**14. Partnerships are vital for responsible AI (SDG 17).**

① Strongly Disagree ② Disagree ③ Neutral ④ Agree ⑤ Strongly Agree

**15. I would join AI governance forums.**

① Strongly Disagree ② Disagree ③ Neutral ④ Agree ⑤ Strongly Agree

# Section E: Sustainability and Risk Literacy (SDG 6–15)

**16. I understand how AI monitors climate impacts**.

① Strongly Disagree ② Disagree ③ Neutral ④ Agree ⑤ Strongly Agree

**17. AI should minimize energy footprint.**

① Strongly Disagree ② Disagree ③ Neutral ④ Agree ⑤ Strongly Agree

**18. AI supports responsible production (SDG 12).**

① Strongly Disagree ② Disagree ③ Neutral ④ Agree ⑤ Strongly Agree

**19. I am aware of AI's environmental risks.**

① Strongly Disagree ② Disagree ③ Neutral ④ Agree ⑤ Strongly Agree

**20. AI education should include climate content.**

① Strongly Disagree ② Disagree ③ Neutral ④ Agree ⑤ Strongly Agree

# Section F: AI–SDG Linkages (Nexus Awareness)

**21. AI can positively contribute to all SDGs.**

① Strongly Disagree ② Disagree ③ Neutral ④ Agree ⑤ Strongly Agree

**22. Lack of AI literacy slows SDG progress.**

① Strongly Disagree ② Disagree ③ Neutral ④ Agree ⑤ Strongly Agree

**23. Select the Top 3 SDGs benefiting from ethical AI use.**

| | | |
|---|---|---|
| 1 | **No Poverty (SDG 1)** – Using AI for social-protection targeting and inclusive welfare systems | ☐ |
| 2 | **Zero Hunger (SDG 2)** – Precision agriculture and smart food-supply chains | ☐ |
| 3 | **Good Health and Well-Being (SDG 3)** – Medical imaging, diagnostics, and disease prediction | ☐ |
| 4 | **Quality Education (SDG 4)** – AI-assisted learning personalization and accessibility | ☐ |
| 5 | **Gender Equality (SDG 5)** – Reducing bias and supporting inclusion in workplaces | ☐ |
| 6 | **Clean Water and Sanitation (SDG 6)** – Monitoring water quality and infrastructure | ☐ |
| 7 | **Affordable and Clean Energy (SDG 7)** – Energy-efficiency optimization and smart grids | ☐ |
| 8 | **Decent Work and Economic Growth (SDG 8)** – Automation ethics and workforce reskilling | ☐ |
| 9 | **Industry, Innovation and Infrastructure (SDG 9)** – Sustainable manufacturing and innovation | ☐ |
| 10 | **Reduced Inequalities (SDG 10)** – Fair-access algorithms and inclusive policy support | ☐ |
| 11 | **Sustainable Cities and Communities (SDG 11)** – Smart-city management and transport | ☐ |
| 12 | **Responsible Consumption and Production (SDG 12)** – Waste-tracking and efficiency | ☐ |

| 13 | **Climate Action (SDG 13)** – Environmental prediction, carbon tracking | ☐ |
|---|---|---|
| 14 | **Life Below Water (SDG 14)** – Marine-ecosystem monitoring | ☐ |
| 15 | **Life on Land (SDG 15)** – Biodiversity monitoring, reforestation analytics | ☐ |
| 16 | **Peace, Justice and Strong Institutions (SDG 16)** – Transparency and anti-corruption tools | ☐ |
| 17 | **Partnerships for the Goals (SDG 17)** – AI collaboration networks and open data sharing | ☐ |

**24. AI literacy should be recognized as cross-cutting enabler.**
① Strongly Disagree ② Disagree ③ Neutral ④ Agree ⑤ Strongly Agree
**25. I feel responsible to promote ethical AI use.**
① Strongly Disagree ② Disagree ③ Neutral ④ Agree ⑤ Strongly Agree

# Section G: Barriers and Enablers

**26. Which factors limit AI literacy development in your environment? (Select all that apply)**
◻ Lack of training
◻ Policy gaps
◻ Ethical mistrust
◻ Low institutional support
◻ High cost
◻ Other: ___________

**27. What actions would most improve AI literacy for sustainability? (Select up to 3)**
◻ Government policy
◻ Curriculum training
◻ Industry partnerships
◻ Public awareness
◻ Open data
◻ Innovation labs

**28. Please share one example of how AI could support an SDG in your context.**
_______________________________________________

### A.10 Scoring and Analytical Framework

| Domain | Question Range | Computation Formula | Interpretation / Purpose |
|---|---|---|---|
| **Technical Literacy (TL)** | Q1–Q5 | Mean(TL) | Cognitive and functional AI understanding |
| **Ethical Literacy (EL)** | Q6–Q10 | Mean(EL) | Awareness of fairness, privacy, and ethical issues |
| **Governance Literacy (GL)** | Q11–Q15 | Mean(GL) | Civic engagement and policy knowledge |
| **Sustainability Literacy (SL)** | Q16–Q20 | Mean(SL) | Awareness of environmental and resource implications of AI |
| **Nexus Awareness Index (NAI)** | Q21–Q25 | Mean(NAI) | Holistic perception of AI's role across SDGs |

### A.11 Notes for Administration

- **Target Group:** University students, educators, researchers, and professionals across sectors.
- **Estimated Completion Time:** 8–10 minutes.
- **Delivery Mode:** Google Form: https://forms.gle/WWspZULYeGhy2xMi7
- **Data Analysis:** Descriptive statistics, correlation, and composite index scoring for TL, EL, GL, SL, and NAI.